\documentclass[pdflatex,sn-nature]{sn-jnl}% Style for submissions to Nature Portfolio journals
\usepackage{graphicx}%
\usepackage{multirow}%
\usepackage{amsmath,amssymb,amsfonts}%
\usepackage{amsthm}%
\usepackage{mathrsfs}%
\usepackage[title]{appendix}%
\usepackage{xcolor}%
\usepackage{textcomp}%
\usepackage{manyfoot}%
\usepackage{booktabs}%
\usepackage{algorithm}%
\usepackage{algorithmicx}%
\usepackage{algpseudocode}%
\usepackage{listings}%
\usepackage{makecell}
\usepackage{xfp}
\usepackage{listings} %for line break in verbatim

\usepackage{lineno}
\usepackage{marginnote}

\newcounter{wordcounter}

\newcommand{\wc}[1]{%
  \addtocounter{wordcounter}{#1}%
  \marginnote{\small\thewordcounter}%
}

\renewcommand{\wc}[1]{}
\makeatletter
\newcommand*{\centerfloat}{%
  \parindent \z@
  \leftskip \z@ \@plus 1fil \@minus \textwidth
  \rightskip\leftskip
  \parfillskip \z@skip}
\makeatother

\theoremstyle{thmstyleone}%
\theoremstyle{thmstyletwo}%

\theoremstyle{thmstylethree}%

\usepackage{enumerate}
\usepackage{enumitem}
\usepackage{amsmath,amsfonts,bm}

\def\eqref#1{equation~\ref{#1}}
\def\1{\bm{1}}

\def\vb{{\bm{b}}}

\def\vh{{\bm{h}}}

\def\vl{{\bm{l}}}

\def\vp{{\bm{p}}}
\def\vq{{\bm{q}}}

\def\vs{{\bm{s}}}

\def\vv{{\bm{v}}}

\def\vy{{\bm{y}}}

\def\mW{{\bm{W}}}

\DeclareMathAlphabet{\mathsfit}{\encodingdefault}{\sfdefault}{m}{sl}
\SetMathAlphabet{\mathsfit}{bold}{\encodingdefault}{\sfdefault}{bx}{n}

\newcommand{\SoneNBadOpenqSubs}{61}
\newcommand{\SoneNall}{770}
\newcommand{\SoneNanGadSubs}{2}
\newcommand{\SoneNanPhqSubs}{2}
\newcommand{\SoneNanSdsSubs}{2}
\newcommand{\SoneNfinal}{704}
\newcommand{\SoneOverallScoreBiasCoeff}{b=-0.58, t(5608)=-24.62, p$<$0.001}
\newcommand{\SoneOverallScoreBiasIntercept}{b=1.01, t(5608)=25.19, p$<$0.001}
\newcommand{\SsaeNfinal}{706}

\newcommand{\SthreeDemandMood}{b=0.02; t(257)=0.32, p=0.748}
\newcommand{\SthreeDemandPhq}{b=-0.07; t(548)=-1.80, p=0.073}
\newcommand{\SthreeDemandRecall}{b=-0.01; t(548)=-0.12, p=0.903}
\newcommand{\SthreeDemandSsae}{b=0.22; t(547)=0.48, p=0.634}
\newcommand{\SthreeIntDiarySsae}{b=2.41, t(549)=21.12, p$<$0.001}
\newcommand{\SthreeMoodChange}{b=-0.10, t(259)=-7.28, p$<$0.001}
\newcommand{\SthreeMoodVsIntsim}{b=0.26, t(257)=2.13, p=0.034}
\newcommand{\SthreeMoodVsSsae}{b=-0.03, t(258)=-6.60, p$<$0.001}
\newcommand{\SthreePhqqtwoChange}{b=0.02, t(550)=1.98, p=0.048}
\newcommand{\SthreePhqqtwoVsIntsim}{b=0.15, t(548)=2.05, p=0.041}
\newcommand{\SthreePhqqtwoVsSsae}{b=0.01, t(549)=3.40, p$<$0.001}
\newcommand{\SthreeRecallChange}{b=-0.11, t(550)=-4.68, p$<$0.001}
\newcommand{\SthreeRecallVsIntsim}{b=0.31, t(548)=1.59, p=0.113}
\newcommand{\SthreeRecallVsSsae}{b=-0.02, t(549)=-3.76, p$<$0.001}
\newcommand{\SthreeSsaeQtwofourChange}{b=0.15, t(549)=1.43, p=0.152}
\newcommand{\SthreeSsaeQtwofourIntercept}{b=-0.39, t(549)=-2.91, p=0.004}
\newcommand{\StwoBadRecallSub}{10}
\newcommand{\StwoBadRecallsentenceSub}{3}
\newcommand{\StwoBadSubInt}{2}
\newcommand{\StwoBadSubOq}{6}
\newcommand{\StwoBadSubOqManual}{1}
\newcommand{\StwoBadSubRec}{5}
\newcommand{\StwoFinalAutobiofalseMh}{146}
\newcommand{\StwoFinalAutobiofalseMl}{139}
\newcommand{\StwoFinalAutobiotrueMh}{133}
\newcommand{\StwoFinalAutobiotrueMl}{135}
\newcommand{\StwoFinalMh}{279}
\newcommand{\StwoFinalMl}{274}
\newcommand{\StwoFinaln}{553}
\newcommand{\StwoIncompleteMoodSub}{8}
\newcommand{\StwoIncompletePhqSub}{15}
\newcommand{\StwoOutlierMoodSub}{6}
\newcommand{\StwoOutlierPhqSub}{10}
\newcommand{\StwoRawAutobiofalseMh}{159}
\newcommand{\StwoRawAutobiofalseMl}{154}
\newcommand{\StwoRawAutobiotrueMh}{143}
\newcommand{\StwoRawAutobiotrueMl}{151}
\newcommand{\StwoRawMh}{302}
\newcommand{\StwoRawMl}{305}
\newcommand{\TableDemographics}{
			Sample size (n) & 704 & 146 & 139 & 133 & 135 \\
			Age (mean ± SD) & 36.8 ± 12.2 & 37.6 ± 11.9 & 36.9 ± 10.7 & 33.6 ± 10.3 & 34.3 ± 9.7 \\
			Sex (\% Female) & 58\% F & 52\% F & 51\% F & 67\% F & 67\% F \\
			\hline
			Ethnicity \% &  &  &  &  &  \\
			\quad White & 81.6 & 89.0 & 88.5 & 91.7 & 89.6 \\
			\quad Black & 11.4 & 1.4 & 2.9 & 0.8 & 0.7 \\
			\quad Mixed & 4.1 & 6.8 & 7.2 & 3.0 & 5.2 \\
			\quad Asian & 2.6 & 2.1 & 0.7 & 4.5 & 3.7 \\
			\quad Other & 0.3 & 0.7 & 0.7 & - & 0.7 \\
			\hline
			Employment \% &  &  &  &  &  \\
			\quad Full-Time & 56.5 & 54.6 & 45.8 & 51.6 & 47.1 \\
			\quad Part-Time & 19.3 & 19.2 & 26.7 & 22.6 & 23.5 \\
			\quad Unpaid & 13.1 & 14.6 & 16.7 & 12.1 & 12.6 \\
			\quad Other & 3.7 & 3.1 & 2.5 & 4.8 & 3.4 \\
			\quad Unemployed & 6.6 & 6.9 & 7.5 & 6.5 & 11.8 \\
			\quad Starting soon & 0.8 & 1.5 & 0.8 & 2.4 & 1.7 \\
			\hline
			Total score (mean ± SD) &  &  &  &  &  \\
			PHQ-9 (depression) & 11.4 ± 7.0 & 11.2 ± 5.8 & 12.5 ± 5.1 & 12.1 ± 5.7 & 12.9 ± 4.9 \\
			GAD-7 (anxiety) & 9.0 ± 6.1 & - & - & - & - \\
			SDS (depression) & 48.4 ± 12.2 & - & - & - & - \\
}
\def\rethicsInfo{The study was approved by UCL's Research Ethics Committee (Project ID: 16639/001)}

\begin{document}

% \title[Article Title]{Internal narratives parameterise affective states}
% \title[Article Title]{Metareasoning constraints determine interactions between narratives, affect and cognition}
\title[Article Title]{Metareasoning constraints couple narratives, affect and cognition}


\author*[1]{\fnm{Jakub} \sur{Onysk}}\email{jakub.onysk.22@ucl.ac.uk}
\author[1,2]{\fnm{Quentin J. M.} \sur{Huys}}\email{q.huys@ucl.ac.uk}

\affil[1]{Applied Computational Psychiatry Lab, Max Planck UCL Centre for
Computational Psychiatry and Ageing Research, Department of Imaging
Neuroscience, Queen Square Institute of Neurology and Mental Health Neuroscience Department, Division of Psychiatry, University College London}

% \affil*[1]{\orgdiv{Department}, \orgname{Organization}, \orgaddress{\street{Street}, \city{City}, \postcode{100190}, \state{State}, \country{Country}}}

% \affil[2]{\orgdiv{Department}, \orgname{Organization}, \orgaddress{\street{Street}, \city{City}, \postcode{10587}, \state{State}, \country{Country}}}

% \affil[3]{\orgdiv{Department}, \orgname{Organization}, \orgaddress{\street{Street}, \city{City}, \postcode{610101}, \state{State}, \country{Country}}}


\abstract{
Narratives and emotions shape thoughts, and thoughts shape our feelings and stories we tell. Why narrative, affective and cognitive states interact remains
unclear. We examine whether this mutual relationship reflects constraints on metareasoning – deciding what to think about – imposed by a shared computational state.
Combining self-report and quantification of depression narratives using large language
models, Study 1 (n=\fpeval{\SoneNfinal}) shows narrative state structure
closely reflects the factorial structure in formal affect assessments, and that
perturbation of the narrative state has commensurate effects on affect via a
latent computational state. 
Study 2 (n=\fpeval{\StwoFinaln}) uses exposure to structured narratives to test
model predictions causally {\em in vivo}. Narrative exposure has consistent
effect on narrative states, with consequences on momentary mood,
cognition, and affect. Critically, effects are predicted by latent
computational state engagement.
This supports the hypothesis that metareasoning constraints determine interactions between narratives, cognition and affect via a shared computational state.\\

\parbox{.3\textwidth}{\small
{\bf Word count}
\\Abstract: 150/150
\\Total: 4177
\\Figures \& Tables: 5 
\\References: 115
}
}

% \fpeval{\SoneNfinal+\StwoFinaln}


\maketitle
\clearpage

\section{Introduction}

Affective states such as moods and emotions influence decisions, thoughts and
cognition. Conversely, thoughts and decisions also influence affective states.
The interaction between cognition and affective states is strongly reflected in
internal narratives, and is sensitive to language. Happiness promotes narratives
such as "I was great today", while hopeless feelings might constrain the space
of thoughts to ones such as: "Life is not worth living". Crucially, thoughts
are targeted by psychological interventions through language to treat affective
disorders. Here, we experimentally examine one potential account for the close
linkage between affect, narrative and cognition: metareasoning – the
internal decision to prioritise certain evaluations over others. 

Several facets of the mutual constraint between affective states, cognition and
narratives have been described.
First, the impact of affective states on cognition is well-established
\cite{LernerKassam15,andrews-hanna_conceptual_2022, raffaelli_think_2021,
bellana_narrative_2022}. Negative emotions in anxiety or depression bias
attention towards specific features in the environment
\cite{matthews_cognitive_1999,schupp_emotion_2006}. Emotions shape
interpretations \cite{siegel_emotion_2018,niedenthal_perception_1991} and
decision-making \cite{Loewenstein96,
	seymour_emotion_2008,LernerKassam15,rutledge_computational_2014,eldar_mood_2016,
emanuel_emotions_2023}, and mood plays a major role in biasing memory, often
manifesting as mood-congruent bias in encoding and retrieving memories
\cite{ellis_mood_1999}. 
Second, cognition also shapes affect. This is evident in cognitive
psychotherapies, where the use of language to change thoughts and
interpretations and to challenge predictions forms the core of the intervention
to change the subsequent affective state \cite{Beck87}. Altering
the consideration set from "hurting someone" to "gifting them a present" has
immediate influence on the subjective emotional state of anger \cite{Linehan93}.
Third, language and narratives are central to mental state assessment, including
when measuring self-reported symptoms to reveal the underlying affective state
\cite{WatsonTellegen88, kroenke_phq-9_2001, hur_language_2024, WulffMata25,
seabrook_predicting_2018}.  Language use is altered in affective
\cite{wang_depression_2013, raffaelli_think_2021, seabrook_predicting_2018},
anxiety \cite{low_natural_2020} and psychotic \cite{nour_trajectories_2023}
disorders. Furthermore, the semantic structure between concepts shapes thinking
tendencies, including suicidal ideation \cite{nock_measuring_2010}. 

The nature of and reason for the three-way interaction between affect, internal
narratives and cognition is, however, poorly understood. One possibility is that
this joint constraint reflects a prioritisation of computations. Prioritisation
of computations is key to the functioning of the human brain. At any one point
in time, the brain could consider a vast array of perfectly unhelpful options
such as the relative shades of our toenails two weeks ago.  In healthy states,
we tend not to get distracted by consideration of irrelevant options, but
instead prioritise evaluations in a manner broadly consistent with our goals.
Solving this metareasoning problem perfectly – deciding what to think about and how much – 
is radically intractable and mandates approximations
\citep{HayRussell11,russell_principles_1991,huys_bonsai_2012, huys_formal_2017,
erdman_computational_2023, trier_emotions_2025}. Fast and frugal narrowing of
consideration sets likely guides flexible, though sometimes maladaptive \citep{andrews-hanna_conceptual_2022},
behaviour. Multiple approximations have been described in decision-making
\citep{huys_bonsai_2012,lieder_anchoring_2018,vanOpheusdenMa23,collins_people_2026,ho_people_2022}, which highlight considerable sensitivity of employed strategies to time pressure, limited cognitive resource and punishments.
Likewise, a narrowing of consideration sets in keeping with maintenance of a narrative
structure is core to language function, and loosening of this constraint a
hallmark of thought disorder \citep{nour_trajectories_2023}, while affective states similarly
strongly constrain consideration sets, e.g. on flight or approach
\citep{huys_formal_2017}. The efficacy of psychotherapeutic interventions such
as acting against the emotion \cite{Linehan93} and metaphors
\citep{stott2010oxford} show that altering consideration sets alters
affective states. Books, stories and movies use narratives to successfully
control and alter consideration sets and affective states. Overall, this
suggests that, akin to decision-making, the mutual constraint between affect, cognition and narratives
might reflect a shared constraint on consideration sets as an approximate
solution to the metareasoning problem \citep{onysk_computational_2026}.
Alteration of this shared state could then explain the mutual influence between
affective, cognitive and narrative states. 

Here, we report the first test of specific quantitative predictions of this
suggestion in two large human samples involving over 1000 participants. First,
we examine the relationship between internal narrative states and formal
assessments of affect relevant to mental disorders. For internal narratives and
affective states to relate to the same underlying state, it must be possible to
derive quantitative affective state measures from samples of internal
narratives. Next, the affective state measures derived from narratives must
respect the established factor structure of the validated instruments (e.g.
\citealp{romera_factor_2008,kroenke_phq-9_2001,spitzer_brief_2006}).  Finally, constraining the consideration set with
narratives must constrain the affective state in a commensurate manner.  Specifically,
causally altering narrative states should alter cognitive measures, and should
also affect relevant cognitive functions. Furthermore, the strength of causal
effects should be proportional to the engagement of the underlying state. Our
approach leverages quantification of affective states through use of validated
instruments of depression (PHQ-9; \citealp{kroenke_phq-9_2001} and SDS;
\citealp{zung_self-rating_1965}) and anxiety (GAD-7; \citealp{spitzer_brief_2006}), as well as protocols based on mood induction \cite{cowen_self-report_2017,
gross_emotion_1995, joseph_manipulation_2020,westermann_relative_1996,
fernandez-perez_use_2022,baker_effects_1993}. Narrative states are quantified
through large language models (LLMs), whose context effects may capture or
parallel some of the constraints considered here \cite{onysk_computational_2026}.
Indeed, LLMs mirror aspects of language and mental states
\cite{colombatto_folk_2024, dillion_can_2023, van_duijn_theory_2023,
hagendorff_machine_2024, shanahan_role_2023}, language processing and generation
\cite{tuckute_language_2024, mahowald_dissociating_2024}, and correlate with
neural representations \cite{goldstein_shared_2022,
tikochinski_incremental_2025, tang_semantic_2023, tuckute_driving_2024}. Finally,
cognition is quantified through use of established memory tasks
\citep{blaney_affect_1986}. 

The results provide key support for the notion that a computational state
characterised by constraints on evaluation sets underlies the mutual influences
between narrative, cognitive and affective states.  The results also inform
broader considerations for mental state assessment and interventions.
\wc{801}

\section{Results}

Across two studies, we recruited $\SoneNfinal$ and $\StwoFinaln$ fluent English
speakers with diverse, on average moderate depression (see recruitment
	Methods~§\ref{methods:rec}, and demographics in Supplementary
	Table~\ref{apdx:demo_table} and
Fig.~S\ref{fig:apdx_s1_totals}-\ref{fig:apdx_s2_totals}).\wc{26}

\subsection{Sampling and quantifying internal narratives}

\begin{figure}[!hbt]
    \centering
    \centerfloat
    \includegraphics{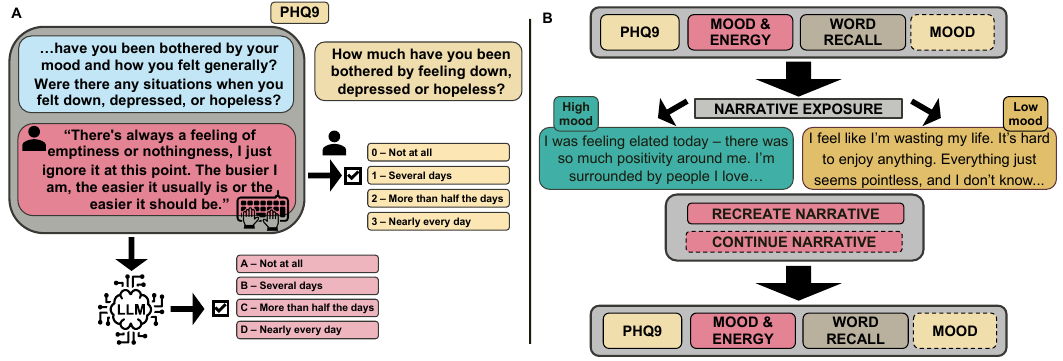}
    \caption{\textbf{A: Narrative sampling Study 1.} We asked participants eight open-ended depression symptom questions derived from PHQ-9 depression questionnaire. Participants provided free-text symptom narratives and then completed multiple-choice PHQ-9, GAD-7 and SDS questionnaires, assessing depression and anxiety. We then created LLM prompts based on participants open-ended symptom narratives to predict the multiple-choice labels. Example question and responses assessing PHQ-9 "Feeling down, depressed, or hopeless" symptoms are shown. 
    \textbf{B: Narrative exposure Study 2.} We measured participants baseline depression symptoms (PHQ-9 scores and open-ended narratives about mood and energy levels), emotional word recall and momentary mood. Participants then engaged with one of positive or negative narrative exposure conditions. Specifically, they listened to emotional narratives that exemplified one of the "Feeling down, depressed, or hopeless" symptom extrema from Study 1. Participants first recreated what they heard to then create new narrative continuations matching the original ones as closely as possible. One group of participants was asked to reflect about similar situations from their own life as a seed for the diary continuation. We repeated the affective measures (momentary mood, emotional word recall, PHQ-9), but instead asked participants for positive re-evaluation of their baseline open-ended narratives about mood and energy levels.}
    \label{fig:study_des}
\end{figure}

Study 1 (n=$\SoneNfinal$) examined the relationship between narrative and
affective states (Fig. \ref{fig:study_des}A).  Narrative states were sampled by
asking participants to provide free text narrative responses to eight open-ended
questions derived from the first eight questions of the Patient Health
Questionnaire (PHQ-9; \citealp{kroenke_phq-9_2001, hur_language_2024}; see
supplementary information \ref{apdx:open_qs} for the exact phrasing of the
questions).  Participants had 1.5 minutes to answer each open-ended question
with a written response of at least 30 words.  Following the open-ended
questions, participants completed several standard validated, quantitative
measures of affect, namely the PHQ-9 questionnaire, and two additional
questionnaires assessing anxiety (Generalised Anxiety Disorder (GAD-7)
\citealp{spitzer_brief_2006}) and depression (Self-rating Depression Scale
(SDS); \citealp{zung_self-rating_1965}).\wc{103}

If participants' symptom narratives indeed arise from an underlying state shared
with the affective state as quantified by the validated instruments, it must be
possible to quantitatively predict self-reported affect measures from the
narrative samples.  We assessed this necessary condition by using LLMs
(\citealp{gemma_2024,llama3modelcard,lian2023mistralorca1}; see Methods
§\ref{appdx:llm_dets}). We created LLM prompts which consisted of the open-ended
symptom question, participants' open-ended responses, and sampled multiple
choice responses from a distribution restricted to valid questionnaire responses
(full example prompt in Supplementary Information \ref{apdx:item_prompt}). This
was done using off-the-shelf LLMs without any finetuning. \wc{87}

\subsection{Internal narrative samples quantitatively recover underlying affective states}

\begin{figure}[!htb]
    \centering
    \centerfloat
    \includegraphics{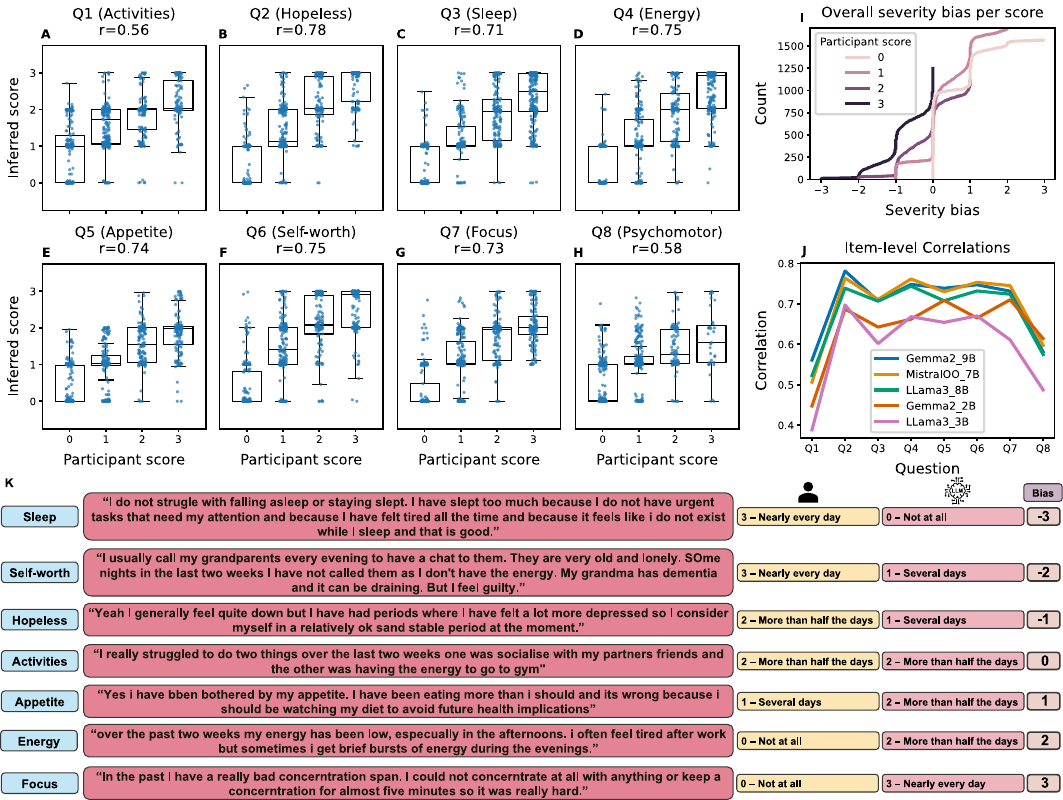}
    \caption{\textbf{Affective score inference}. For each open-ended question (\textbf{A-H}), we plot the best performing model's (Gemma2-9B) average item-level inferred scores against the ground-truth from each participant. Across all questions, the model predicted the scores well – correlation: $0.58-0.78$ (p$<0.001$). \textbf{I}: We quantified the symptom severity bias as model's deviation from the ground-truth. For each score (0-3), we plot the cumulative count of the biases, and observe a systematic overestimation of scores for no symptom response and an underestimation as the symptom severity increases (see Supplementary Information \ref{adpx:itemlogit_extra_bias} for question and model specific plots). \textbf{K}: We report example symptom responses for each bias score. \textbf{J}: We report performance (correlation; p $<0.001$) of each LLM at predicting PHQ-8 scores from open-ended QA-pairs, see scatter plots for each LLM in Supplementary Information \ref{adpx:itemlogit_extra}.}
    \label{fig:study2_llm_item_level_scores}
\end{figure}

Narrative-derived affect
scores for each question (Fig.~\ref{fig:study2_llm_item_level_scores}A-H) were moderately to highly correlated with the true
affect scores ($0.56-0.78$). The results are near the limit of resolution
achievable with the questionnaires, given the established test-retest
reliability of the PHQ-9 instrument of around 0.84 over 48-hours
\cite{kroenke_phq-9_2001}, and two-week item-level reliabilities in the range of
$0.55-0.73$ \cite{sun_reliability_2020}. 
 
Detailed examination of individual items revealed informative systematic biases
in the narrative-derived LLM affect scores.
Fig.~\ref{fig:study2_llm_item_level_scores}I shows that narrative-derived LLM
scores across all questions were biased, with the response for no symptoms being
systematically overestimated (\SoneOverallScoreBiasIntercept), and then
underestimated as the severity increased (\SoneOverallScoreBiasCoeff).
Furthermore, we see lower performance for Q1 ("Activities") and Q8
("Psychomotor").  This pattern of performance across questions is similar across
different LLMs , with smaller (3B-parameter) model generally performing worse
(Fig.~\ref{fig:study2_llm_item_level_scores}J).
These biases reflected at least two processes. At the extremes, they reflected
inconsistencies in participant data, where participants' narrative samples
contained evidence of symptoms, while their scores indicated absence, or vice
versa.  Fig~\ref{fig:study2_llm_item_level_scores}K shows illustrative examples.
Similar deviations are present at item-level (Supplementary Information
\ref{adpx:itemlogit_extra_bias}). In terms of items Q1 and Q8, the biases also
reflected details of the narrative probe. For these questions, the probe was
"Can/could you [share/describe]...?", whereas for the remaining questions the
probes were bipartite.

These results establish that narrative states are in keeping with mental
affective state quantifications through validated instruments, and that there is
high sensitivity to the specifics of the narrative probe. These data, combining
samples from internal narratives and self-report assessments are likely
informative about the upper boundary on realistically achievable measurements
derived from text that is less directly related to indviduals' internal state,
e.g. text sampled from care providers in electronic health records.\wc{280}

\subsection{Internal narratives reveal consistent computational structure}

\begin{figure}[!hbt]
    \centering
    \centerfloat
    \includegraphics{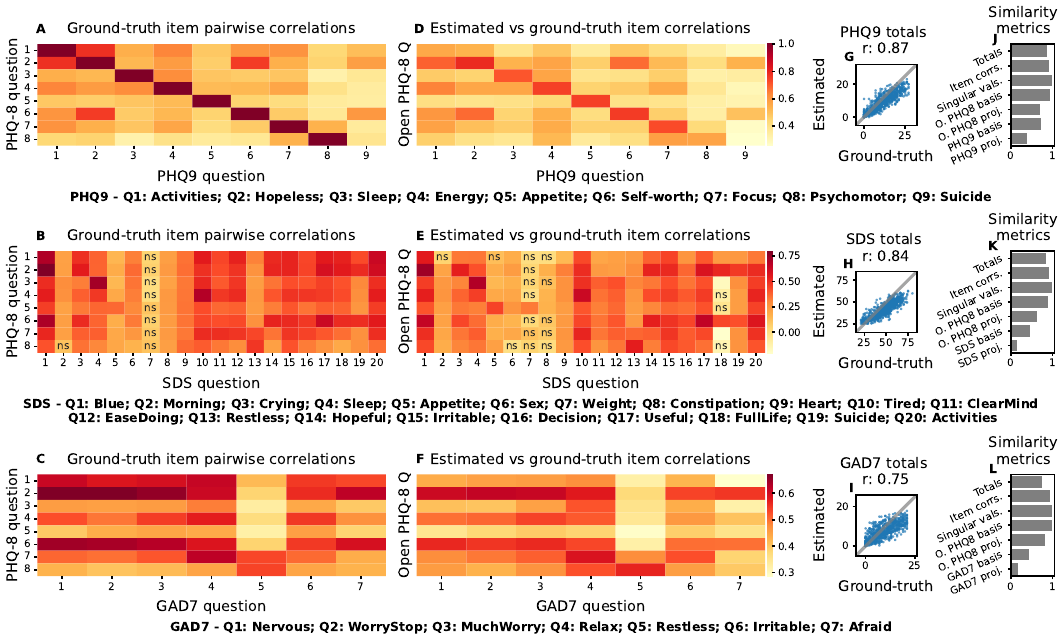}
       \caption{\textbf{Computational structure of affect}. 
       \textbf{A-C}: Participants’ ground-truth pairwise correlations between PHQ-8 item scores and PHQ-9 (\textbf{A}), SDS (\textbf{B}) and GAD-7 (\textbf{C}) scores (ns - non-significant correlations). \textbf{D-F}: For each open-ended PHQ-8 question (y-axis), Gemma2\_9B inferred PHQ-9 (\textbf{D}), SDS (\textbf{E}) and GAD-7 (\textbf{F}) item scores (x-axis) comparably to the ground-truth. \textbf{G-I}: Model total score estimate captured the total ground-truth severity for PHQ-9 (\textbf{G}), SDS (\textbf{H}) and GAD-7 (\textbf{I}) questionnaire. \textbf{J-L}: We quantified the closeness-of-fit between inferred and ground-truth questionnaire structures for PHQ-9 (\textbf{J}), SDS (\textbf{K}) and GAD-7 (\textbf{L}) (values closer to 1 indicate a more similar response structure) – total scores correlations, average item-level correlations difference complement, singular value decomposition bases and projection measures. Results for other LLMs are reported in Appendix \ref{adpx:gen_logit_extra}.
    }
    \label{fig:study2_llm_gen_item_qs}
\end{figure}

If narratives states reflect the same constraints as the affective state, then
narrative state samples must a) respect the factorial structure of established
instruments; b) predict assessments of affective states through related
instruments; and c) predict the factorial structure of other instruments.  As
such, we next characterise the computational structure across narrative samples
and across questionnaires assessing both symptoms of depression, but also
symptoms of anxiety. Anxiety is often comorbid with depression \citep{hirschfeld_comorbidity_2001}, and as such
narrative samples about symptoms of depression should contain information about
symptoms of anxiety. Throughout, the ground truth here is given by pairwise item
correlations between questionnaire scores (Fig.~\ref{fig:study2_llm_gen_item_qs}A-C; first column).  These
correlation matrices capture co-occurrence of specific symptoms, and form the
basis for factor analytic identification of the underlying latent structure of
affective states \cite{romera_factor_2008, mulaik_foundations_2013}. For
example, in Fig. \ref{fig:study2_llm_gen_item_qs}B, PHQ-8 question 4 ("Feeling tired or having little energy") correlates
highly with SDS question 10 ("I get tired for no reason") and with question 20
("I still enjoy the things I used to do").
These correlations hence capture a common sub-construct identifiable beyond
details of the specific probe.\wc{183}

To assess this, we constructed LLM quantifications of the narratives using
new prompts. Each prompt consisted of a narrative sample from one participant to
one of the eight open-ended PHQ-8 questions, and each of the multiple-choice questions
from one of the PHQ-9, SDS or GAD-7 questionnaire.
Fig.~\ref{fig:study2_llm_gen_item_qs}D-F (second column) show the estimated correlation
structure between scores, recovered from the narratives. Comparison of the ground truth
(Fig.~\ref{fig:study2_llm_gen_item_qs}A-C) and narrative-derived
correlation matrices (Fig.~\ref{fig:study2_llm_gen_item_qs}D-F)
revealed a visually similar pattern.
\wc{79}

We quantified the visual match through several similarity metrics. Firstly, with
participants' total scores for the PHQ-9 (correlation of 0.87;
Fig. \ref{fig:study2_llm_gen_item_qs}G,J). Secondly, there was a small
generalisation decrement when using narrative samples built around the PHQ-8 to
quantify depression using another instrument (SDS; correlation 0.84;
Fig.~\ref{fig:study2_llm_gen_item_qs}H,K). Third, the narrative responses also
contained substantial information about anxiety (correlation 0.75;
Fig.~\ref{fig:study2_llm_gen_item_qs}I,L). Performance was, in each case,
surpassed by performance at the individual item level (comparing ``Totals'' to
``Item corrs.'' in Fig.~\ref{fig:study2_llm_gen_item_qs}J-L). 

Critically, the narrative sample-derived affect scores reflected co-occurrence
patterns within different sets of symptoms. This was evident across multiple
measures derived from singular value decomposition (SVD) of the correlation
matrices (Fig.~\ref{fig:study2_llm_gen_item_qs}J-L). These measures quantify the match between structures for each
questionnaire, with values closer to 1 showing closer match for that measure
(see Methods §\ref{apdx:cov_match_measures} for details). "Singular val."
captures how closely the LLM-based SVD-estimated structure recovers ground-truth
symptom-grouping scores, while the remaining metrics compare the SVD-based
structures that define the symptom-groupings themselves. Results for
quantification with other LLMs are shown in Appendix \ref{adpx:gen_logit_extra}
and Fig.~\ref{apdx:study1_cov_metrics_all}. 

Overall, narrative samples contained detailed information about affective
structure within and across validated instruments, both within a disorder (SDS)
and across disorder boundaries (GAD-7).
\wc{205}

\subsection{The subspace of internal narratives shows sensitivity to perturbations}
  
The results so far show a close quantitative match between affective and
narrative states. We next attempted to pinpoint the computational state within
LLMs that links narrative to affective states. Critically, this state should
respond to constraints imposed by the narrative sample, and drive the change in
affect. We therefore characterised the underlying structure across internal
narratives by identifying relevant subspaces within LLM hidden states and
established how selectively perturbing individual computational states affects
the subsequent computations.
\wc{77}

\subsubsection{Subspace identification}

\begin{figure}[!hbt]
    \centering
    \centerfloat
    \includegraphics{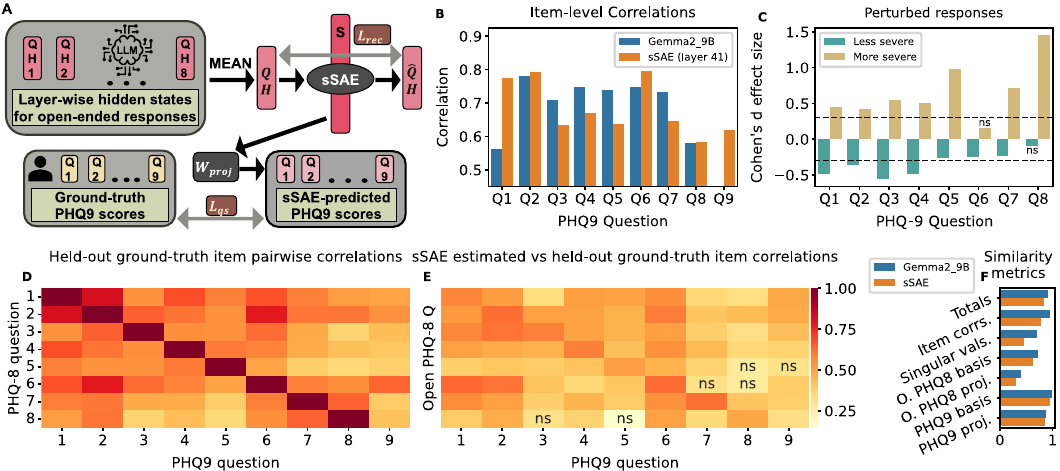}
    \caption{
    \textbf{Supervised sparse auto-encoder (sSAE)}. \textbf{A}: We extracted a layer-wise, average hidden state representation over participants' open-ended narratives and trained an sSAE to predict PHQ-9 scores, reconstructing the average original hidden state.
    \textbf{B}: Example sSAE (best-performing-layer) predicted item-level PHQ-9 scores (orange; p$<0.001$) comparably to the original LLM sampling predictions (blue).
    \textbf{C}: We steered Gemma2-9B responses to each question, given participants' corresponding open-ended narratives by perturbing their hidden state representation. Cohen-d effect size reveals significant perturbation towards more (yellow) and less (turquoise) PHQ-9 symptom severe direction. Dashed lines indicate small effect size cut-off $|d|<0.3$, ns marks non-significant effects following one-sided t-test.
    \textbf{D}: Held-out participants’ ground-truth pairwise correlations between PHQ-8 item scores and PHQ-9 scores. \textbf{E}: For each open-ended PHQ-8 question (y-axis), best performing sSAE model inferred PHQ-9 item scores (x-axis) comparably to the ground-truth (ns - non-significant correlations). \textbf{F}: We quantified the closeness-of-fit between sSAE inferred and ground-truth PHQ-9 questionnaire structures in orange (values closer to 1 indicate a more similar response structure) – total scores correlations, average item-level correlations difference complement, singular value decomposition bases and projection measures. sSAE metrics (orange) are comparable with previous LLMs sampling approach (blue).}
    \label{fig:s2_ssae_p1}
\end{figure}

Firstly, we identified a symptom-specific subspace across participants'
narratives by analysing LLM representations (hidden state embeddings). We
used the ground truth dataset of symptom verbalisations and scores collected in
Study 1 to train a supervised sparse auto-encoder (sSAE;
\cite{lee_emergence_2025, yun_transformer_2021, le_supervised_2018}) to predict
participants' z-scored PHQ-9 scores from the average hidden state representation
of the open-ended prompts (see Fig.~\ref{fig:s2_ssae_p1}A). This identified a
sparse representations of narratives that correspond to the symptom scores (c.f.
model architecture, training and hyper-parameter tuning in Methods
§\ref{apdx:ssae}). Results below refer to the held-out sample not used for
training the sSAE. \wc{92}

sSAE PHQ-9 item-level prediction correlations for the best LLM layer are shown
in Fig.~\ref{fig:s2_ssae_p1}B, compared to the sampling results from Study 1.
The performance is high, with correlations of 0.6 and higher, with some
variability across items. Importantly, the sSAE-identified subspace still
captured the correlation structure between items. To examine this, we inferred
PHQ-9 scores from individual embeddings of each PHQ-8 narrative sample.
Fig.~\ref{fig:s2_ssae_p1}D and E show the ground-truth structure and the
sSAE-inferred structure, respectively. Reducing the latent is expected to incur
a representative loss, and this is visible. Nevertheless, the similarity metrics
(Fig.~\ref{fig:s2_ssae_p1}F) show that a substantial fraction of the relevant
structure is captured. \wc{108}

\subsubsection{Subspace perturbations}

To ascertain a causal link between sSAE latent representations and the PHQ-9
predictions, we examined the sensitivity and specificity of the subspace to
item-specific perturbations. As a sanity check, we first ensured that we can identify a
perturbation of the
sparse representation that corresponds to specific change in the latent sSAE
PHQ-9 scores (c.f. Methods~§\ref{apdx:ssae_pert};
Fig.~\ref{fig:ssae_apdx_pert}). We then examined the effect of perturbations in
the sSAE subspace on the actual responses in the item-level questionnaire
sampling. We again first embed each open-ended prompt with the LLM, and then
extract its hidden state representations. Using sSAE, we identify the
depression subspace for each prompt and perturb the sparse representation of
each using the method above, both in positive and negative severity direction.
Essentially, we move the original hidden state towards the reconstructed,
perturbed hidden state so as to bias the questionnaire responses (implementation
details in Methods §\ref{apdx:logit_pert}). Fig.~\ref{fig:s2_ssae_p1}C, shows
that most of the responses can be steered, both in the negative (less symptom
severity) and positive (more symptom severity) direction. The sSAE hence
captures the symptom-specific subspace in LLM representations of participants
internal narratives that reflects the underlying ground-truth structure of
validated affective instruments.
\wc{193}

\subsection{Narratives parametrically alter affective and cognitive states}

Narrative samples reflect the structure of validated affective instruments in
substantive detail; and there is a representational subspace which captures how
the constraint due to narrative samples alters affect measures. We next test
this model prediction using causal experimental manipulations. The key question
is whether affective and cognitive states are sensitive to manipulations of the
narrative state, and whether this influence is quantitatively in keeping with
the findings so far, i.e. whether it reflects the same constraints on metareasoning. 
\wc{79}

Study 2 induced targeted narrative states by exposing participants to narratives
and asking them to continue with the narrative (Fig. \ref{fig:study_des}B).
Narratives from Study 1 were combined to generate fictional diaries.
Participants listened to audio version of the narratives, recreated and extended
them. Narrative samples corresponding to high and low PHQ-9 item 2 ("Feeling
down, depressed, or hopeless") scores were chosen for the high and low mood
conditions, respectively (c.f. Supplementary Information \ref{apdx:s3_instr}
and \ref{apdx:s3_proc}). Participants underwent measurements of affective,
narrative and cognitive states before and after the narrative state induction. 

Changes in narrative state serve as a manipulation check. If narrative state
induction results in a consistent restriction of the consideration set, then
processing of the narrative stimuli should move participants towards the
corresponding affective state. We expected a large effect on momentary mood
assessment in keeping with existing mood induction protocols, but also a smaller
effect one the PHQ-9 score, which assesses participants' mood over a period far longer
than the experiment (two weeks versus less than 30 minutes). Furthermore, we also
predicted that the restriction of the consideration set would affect cognition,
specifically biasing recall towards items consistent with the narrative
induction. Critically, all changes should be commensurate with the engagement of
the latent representational state identified using the sSAE model. 
\wc{212}

\subsubsection{Measuring narrative induced computational states}

\begin{figure}[!hbt]
    \centering
    \centerfloat
    \includegraphics{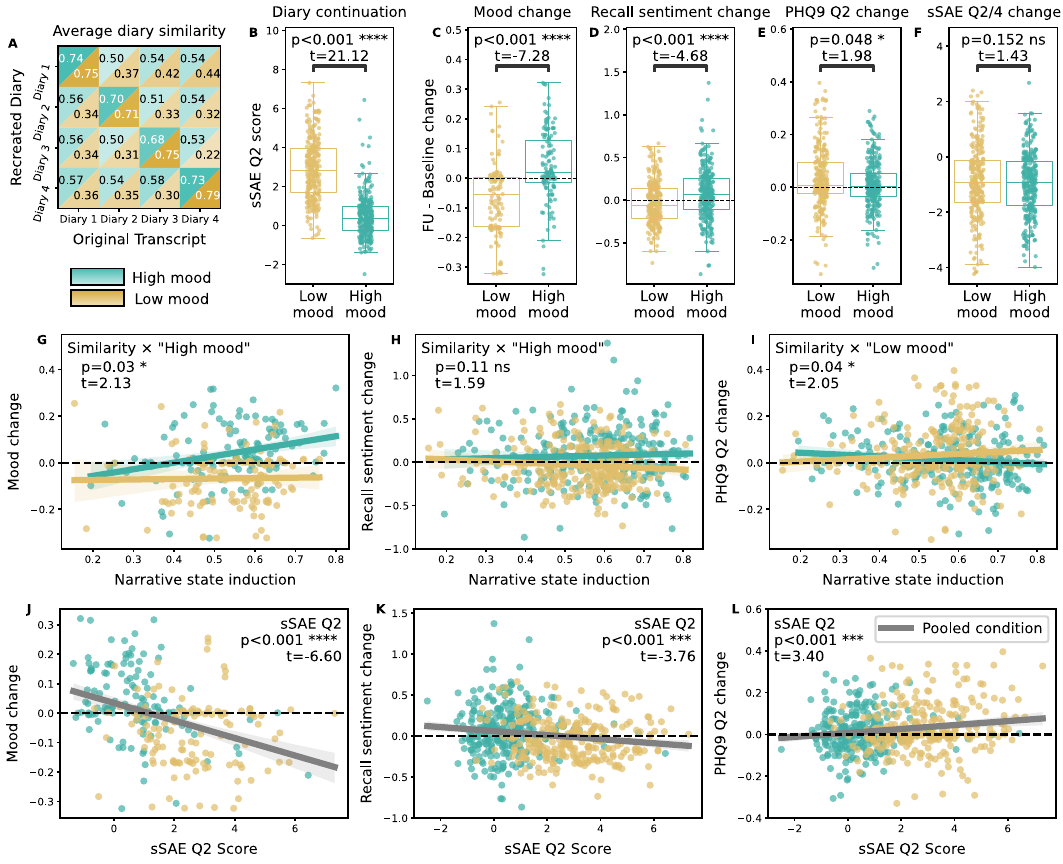}
    \caption{
    \textbf{Narrative exposure Study 2 results} for high mood (turquoise) and low mood (yellow) conditions. \textbf{A}: Average cosine similarity between each of participants’ recreated diary text and the original diary transcript revealed that participants recreated the diaries well.
    \textbf{B}: Supervised sparse auto-encoder Q2 (“Feeling down, depressed, or hopeless”) subspace reflects the positive/negative narrative state engagement across participant diary continuations.
    \textbf{C-E}: Affective and cognitive measure changes after mood induction (momentary
	 mood, word recall sentiment, PHQ-9 Q2 score; follow-up [FU]) differ between
	 conditions, in keeping with narrative exposure.
    \textbf{F}: Average mood-energy (Q2/Q4) sSAE score change between baseline open-ended mood and energy symptoms responses and FU symptom positive re-evaluation reveal successful positive re-framing of internal narratives.
	 \textbf{G-I}: Moderation analysis of the change measures (momentary mood, word recall sentiment, PHQ-9 Q2 score) 
     reveals an effect of the extent of narrative state induction on affective and cognitive changes.
    \textbf{J-L}: Affective and cognitive measures change (momentary mood, word recall sentiment, PHQ-9 Q2 score), pooled across conditions, reflect the latent sSAE Q2 computational state consistency. All statistical results are from linear regression analyses predicting each measure (y-axis) as a function of condition (and the x-axis moderator variable in G-L), controlling for baseline total PHQ-9 score.}
    \label{fig:mood_ind}
\end{figure}

Participants recreated diaries matched each of the original diaries well in
terms of textual similarity (Fig.~\ref{fig:mood_ind}A) and in terms of sSAE
subspace engagement (Fig.~\ref{fig:mood_ind}B; \SthreeIntDiarySsae).  As
hypothesised, the narrative state intervention had a strong effect for the most
proximal, momentary mood measure (Fig.~\ref{fig:mood_ind}C; \SthreeMoodChange).
The narrative intervention also altered cognition, leading to a negative
affective recall bias (Fig~\ref{fig:mood_ind}D; \SthreeRecallChange).
Critically, participants' ratings of their PHQ-9 symptom severity of "Feeling
down, depressed, or hopeless over the past two weeks" also changed in accordance
with narrative state manipulation (Fig~\ref{fig:mood_ind}E;
\SthreePhqqtwoChange), with participants' severity rating lower/higher following
High/Low mood narrative exposure. The average latent sSAE scores of the mood and
energy open-ended narrative samples did not differentiate between the two groups
(Fig.~\ref{fig:mood_ind}F; \SthreeSsaeQtwofourChange), with both groups
re-evaluating their symptoms in the positive direction
(\SthreeSsaeQtwofourIntercept).  Controlling for participants' demand-effect
judgements did not alter results on momentary mood (\SthreeDemandMood), recall
(\SthreeDemandRecall), Q2 symptom score ("Feeling down, depressed, or hopeless";
\SthreeDemandPhq) or latent sSAE Q2/Q4 score (\SthreeDemandSsae). \wc{160}

Importantly, participants whose narrative state was more effectively induced
showed stronger effects. We measured narrative state induction as adherence by
quantifying the similarity in participant diary continuation with the original
diaries. Better narrative induction produced stronger momentary mood change in
the High mood condition (Fig.~\ref{fig:mood_ind}G; \SthreeMoodVsIntsim) and
stronger PHQ-9 Q2 score ("Feeling down, depressed, or hopeless") in the Low mood
condition (Fig.~\ref{fig:mood_ind}I; \SthreePhqqtwoVsIntsim). The effect of the
intervention adherence on the recall bias in the High mood condition was in the
correct direction but failed to reach significance (Fig.~\ref{fig:mood_ind}H;
\SthreeRecallVsIntsim).  Finally, we examined the computational consistency of
the intervention diaries with the latent sSAE Q2 score. This strongly predicted
the extent of mood change (Fig.~\ref{fig:mood_ind}J; \SthreeMoodVsSsae), recall
bias (Fig.~\ref{fig:mood_ind}K; \SthreeRecallVsSsae) and Q2 symptom score
("Feeling down, depressed, or hopeless") change (Fig.~\ref{fig:mood_ind}L;
\SthreePhqqtwoVsSsae).  Overall, narrative induction causally induced changes in
internal state with a coherent set of changes across mood, cognition and
affective state instruments; and the extent of these changes was determined by
computational state consistency.  
\wc{166}

\section{Discussion}\label{sec12}

Cognition, language and emotion share some computational challenges arising from
metareasoning. They all encompass problems that require selection of sequential
combinations amongst a large set of potential candidates. Resource constraints, arising from bounded time or cognition,
imply that only a few of the many options can be evaluated \citep{lieder_resource-rational_2020}. This
in turn implies the need for a decision or prioritisation of options for
evaluation \citep{HayRussell11}. This prioritisation has profound impacts as it
predetermines the potential outcomes of the evaluation process. Affective states
appear to be closely linked to this kind of predetermination \citep{huys_formal_2017},
and shape behaviour, thought, and narratives accordingly: fear prioritises
defensive actions, anger aggressive ones. Conversely, altering the prioritised
set of options alters the affective state \citep{Linehan93,Beck13,WellsEa09}.\wc{115} 

Here, we find support for core predictions in support of this account. The first
study builds on narrative samples to assess the consideration set: concepts
and thoughts expressed in the narrative samples are clearly within the current
consideration set and additionally reflect everyday coupling between thoughts and behaviours. 
These consideration sets contained substantive and
non-trivial quantitative information about the affective state. The affective
state was quantified using standardised and well-established instruments used
for the management of affective disorders such as generalised anxiety disorder
and major depressive disorder. This enabled examination of structural
components, and we found strong evidence that well-established structural
aspects of affect including factor structure, generalisation across instruments
and across disorders are reflected in the contents of narrative samples. These
findings, building on a substantive novel ground-truth dataset, established a
strong quantitative link between narrative samples and affect: what we think
about relates quantitatively to our affective state. \wc{150}

One core argument around affect as addressing metareasoning challenges is that
selection of affective states reduces the computational costs
\citep{huys_formal_2017,onysk_computational_2026}. The problem is reduced from identifying the
optimal consideration set to picking amongst a small set of potential
predetermined consideration sets – those provided by different emotional states.
This reduction predicts that the underlying latent state should be relatively
simple. In the setting of LLM representations of narratives, this suggests that a smaller
subspace should be identifiable, which captures the constraints imposed by the affective state. 
While LLM resource constraints are very different from biological brains, their powerful and rich representations may nevertheless capture those faced by human brains \citep{DeSabbataGriffiths24, binz_foundation_2025,ben-zion_assessing_2025,anthropic_emotion}.
Indeed, the LLM representation subspace we
identified did describe substantive and meaningful aspects of the
narrative state influence on affect, and critically preserved core structural
components. \wc{133}

While Study 1 provided correlational evidence of consistency, Study 2
provided causal evidence. When participants engaged with the narrative samples
from Study 1, they entered a specific narrative state. This narrative state
was characterised by a prioritisation of concepts and language in keeping with
the narrative samples. This prioritisation was then evident not only in the
narrative domain, but also in the affective and cognitive domains: momentary
mood was altered, recall was altered, and even judgements about affect
spanning past two weeks were altered. Critically, the extent to which
this was the case was moderated by engagement of the latent subspace.\wc{101} 

Overall, then, these experiments suggest that affect, cognition and narrative states are
mutually influential because they share reliance on the same, domain-general,
underlying requirement to prioritise evaluation. This highlights the inherently computational nature of affect, mood and emotions and, for example, echoes the view that mood serves as a computational state summarising reward history to guide decision-making \cite{rutledge_computational_2014,eldar_mood_2016, emanuel_emotions_2023}. Our work expands this by quantitatively establishing the role of narratives in shaping computations. The findings further speak to the phenomenon of psychological momentum – a fundamental principle that continually shapes our thoughts, emotions, actions, dispositions to maintain the "momentum" of the state in an adaptive manner \cite{honey_psychological_2023,bellana_narrative_2022}. Crucially, psychological momentum sometimes manifests as maladaptive rumination \cite{andrews-hanna_conceptual_2022} in depression, and our work highlights a putative mechanism for this, where interactions between momentary narrative states bias cognition and self-report, which could potentially contribute to persistent maintenance of symptoms and attitudes. \wc{139}

While do not put forward a unified theory of emotions, our findings and the proposed metareasoning account relate to both appraisal \citep{MoorsEa13,scherer_appraisal_2001} and constructionist \citep{barrett_how_2017,barrett_theory_2017} theory of emotions. As appraisals themselves are computations that the brain must perform, a cognitive state that reduces a consideration set and focuses on a narrow selection of features will likely bias appraisals in specific ways, and, therefore, according to the theory \cite{scherer_appraisal_2001}, influence the emotional state itself. Similarly, the process of narrative exposure biasing memory retrieval facilitates accumulation of evidence to construct narrative-congruent emotional states \cite{barrett_theory_2017} and judgements thereof \citep{serfaty_subjective_2026}. Such inexpensive metareasoning heuristic is also likely adaptive for recurring high stakes situations that leave little time for deliberation \citep{darwin1872expression}. Furthermore, the temporally-persistent nature of emotional strategies may allow for overall adaptive behaviour, exploiting the temporal consistency of the environment where meaningful events unfold over time. \wc{133}

However, an important question remains – how are these computational strategies and biases learned in the first place? While simpler computations could be evolutionarily hard-wired, more complex ones are learned from experience or active deliberation \cite{huys_formal_2017}. Such process of meta-learning effectively amounts to learning different sets of priors over actions across different contexts from experience, forming strategies to guide future learning and behaviour \cite{wang_prefrontal_2018,chen_learning_2026}. For instance, when responding about depression symptoms, a specific learned affective dependency gets more consideration, where instances of low self-worth precede feeling useless and hopeless. 
Perhaps one of the reason that makes LLMs and their representations relevant here, is their ability for in-context learning arising from similarly meta-learned dependences \cite{lampinen_broader_2025,kumar_using_2022, chan_data_2022,han_understanding_2023}. This makes them exceptionally good at rewriting passages in different emotional tones \citep{zhan2024llm,xiao-etal-2024-healme}, providing mental-health support \citep{lim_chatbot-delivered_2022, filienko_toward_2025,
nepal_mindscape_2024, nie_llm-based_2025} and even displaying their own functional hallmarks of emotions \citep{anthropic_emotion}, likely supported by their own workspace of internal computations \citep{anthropic_workspace}. \wc{143}

The implications of affective states as reflecting the prioritisation of
computations are wide-ranging. Conceptually, this framework clarifies how verbal
interventions drive affective change – a crucial insight for psychotherapeutic
'talking therapy' settings. For cognitive science, the approach offers a way to
better quantify the behavioural impact of instructions, particularly benefiting
efforts to measure psychotherapeutic effects using behavioural methods
\citep{NorburyHuys24}. On an applied level, it provides a tool to gauge the
structure of thoughts in depression from free text. Furthermore, its sensitivity
to perturbation could serve as a computational framework for tracking the
outcomes of verbally-rich psychological therapies
\cite{abdou_leveraging_2025,stade_large_2024,brindle_language_2026} and broader
symptom improvement \cite{hur_language_2024}. Ultimately, quantifying this
process of improvement allows us to potentially identify or design {\em in
silico} the most impactful components of psychological change at an individual
level, paving the way for precision psychotherapy and psychiatry.
The work may also be relevant
to AI-human interactions, both good and bad 
\citep{weilnhammer_2026,dohnany_technological_2026,hudon_delusional_2025,siddals_it_2024, hill_they_2025}, as it provides a
causal pathway through which narrative interactions alter and interact with
affect and decisions. Specifically, in the context of widespread reliance on AI for mental health support \citep{cross_use_2024}, it's crucial to evaluate how user's vulnerable cognitive state and intent can deteriorate as a result of AI sycophantic conversational style \citep{weilnhammer_2026}, pushing some towards suicide or delusion \cite{hudon_delusional_2025,dohnany_technological_2026, hill_they_2025}.
\wc{203}

Several limitation of the work should be noted. The data were acquired from
online UK participants. These participants may not fully reflect the wider
population, and may particularly not reflect clinical populations, or the impact
of narrative state manipulations in clinical populations. We focused mainly on
few instruments, and while these were chosen to examine generalisation across
instruments assessing similar (SDS and PHQ-9) and different (GAD-7) disorders,
the wider generalisation structure across different affective states remains to
be studied. Similarly, the generalisation across other cognitive functions
beyond affective recall bias should be examined. The causal work focused on one
affective dimension, and further work would be required to assess the
specificity of the identified latent subspace. Future studies should also
manipulate the meta-reasoning constraints, to causally establish their impact.\wc{129}

This work describes how narrative states both quantitatively reflect and
causally alter affect and cognition through a common latent space.  More simply
put, what we feel, think and say are inextricably bound by a shared
computational mechanism: the fundamental need to prioritise what the brain
evaluates.  These findings offer a unified framework for metareasoning that
bridges cognitive science and clinical practice. By mapping the computational
structure of thought, it opens new avenue for understanding how narratives shape
our inner lives. \wc{80}

\clearpage
\section{Methods}\label{methods}
\subsection{Recruitment}\label{methods:rec}

 \rethicsInfo. All participants (18 or above) provided online consent after reading Participant Information Sheet. Participants were informed that they would be asked questions about their mood and feelings and we have provided information about ways to seek help should they feel affected by the issues raised by these questions. We reimbursed participants at a rate of £8.21/h, which as rate approved by the Ethics Committee (appropriate for the demographic). For the analyses, we have pseudo-anonymised the data (changed the Prolific ID, which in itself ensures participants anonymity), but cannot ensure no identifiable is contained in the open-ended responses. Participants were free to withdraw at any point, and we have emphasised that throughout the study. We recruited a sample of participants from Prolific \cite{noauthor_prolific_nodate} to represent diverse range of depressive symptoms. We screened for participants who were willing to participate in a study about sensitive topics (mental health, emotions, feelings) and potentially harmful content, who had approval rate of 95-100\%, at least 5 previous submissions.

\subsubsection{Study 1 procedure and participants}
In the first study, participants were instructed that they will have to provide written responses as well as multiple-choice responses. They answered one open-ended question at a time within 1.5 minutes, while the multiple-choice questions were displayed on one scrollable page. Additionally, to ensure more genuine responses, we prevented participants from pasting text into the free-text box. Specifically, we included participants who passed quality criteria: 1)Accumulated at most 5 warnings indicating that they timed out or provided less than 30 words for each open-ended question; 2) Failed at most 1 attention check (out of two), where participants had to type specific phrase or select a specific multiple-choice response.
This resulted in recruiting $n=\SoneNall{}$ in total.

Furthermore, during data preprocessing, we excluded participants who:
\begin{itemize}
    \item wrote less than 30 words ($n=\SoneNBadOpenqSubs{}$)
    \item had any of the PHQ9 scores missing ($n=\SoneNanPhqSubs{}$)
    \item had any of the GAD7 scores missing ($n=\SoneNanGadSubs{}$)
    \item had any of the SDS scores missing ($n=\SoneNanSdsSubs{}$)
\end{itemize}

This resulted in the final sample size of $n=\SoneNfinal{}$. We plot the distribution of depression and anxiety severity scores in Fig. \ref{fig:apdx_s1_totals}

\subsubsection{sSAE study participants}
In the supervised sparse auto-encoder study, we started with all recruited participants from Study 1 ($n=\SoneNall{}$) and then excluded participants based on the following criteria:
\begin{itemize}
    \item wrote less than 30 words ($n=\SoneNBadOpenqSubs{}$)
    \item had any of the PHQ9 scores missing ($n=\SoneNanPhqSubs{}$)
\end{itemize}
After exclusion, we ended up with $n=\SsaeNfinal$ for training, validation and test datasets (one was lost due to technical issue when saving their hidden states).

\subsubsection{Study 2 procedure and participants}
In the second study, participants, who reported no dyslexia, speech disorders, hearing difficulties or wearing a cochlear implant, completed the following types tasks:
\begin{itemize}
\item Instructions and practice phase where they had to exactly transcribe "five dogs are drinking water" to ensure the audio is working correctly as well familiarise themselves with the VAS scale.
\item Answer standard nine PHQ9 question on VAS scale. They had 20 seconds to rate each question.
\item Answer two baseline open-ended question about mood and energy levels. They had 100 seconds to write at least 30 words for each.
\item View 18 words and then recall as many words as possible in 45 seconds from the set list ("VIGOROUS", "ENERGETIC", "LIVELY", "EXHAUSTED", "TIRED", "DRAINED", "JOYFUL", "DELIGHTED", "HAPPY", "UNHAPPY", "HOPELESS", "MISERABLE", "GENUINE", "WHOLESOME", "ETHICAL", "CORRUPT", "SLOPPY", "UNSAFE")
\item Rate their momentary mood on the VAS scale ("How happy are you at this moment?”)
\item Listen twice to each of the four diary entries to then recreate each in 40 seconds by typing at least 15 words.
\item Create 100-word long continuation diary entry in 180 seconds, based on the ones they listened and recreated
\item Positively re-evaluate their original open-ended responses about mood and energy in one response of at least 30 ords in 120 seconds.
\end{itemize}

Specifically, we included participants who passed quality criteria: 1) Pass the practice phase technical check; 2) Accumulated at most 5 warnings indicating that they timed out or provided less than required word limit for each open-ended question or the diary task; 3) Failed at most 1 attention check (out of two), where participants had to select a specific multiple-choice response. Additionally, to ensure more genuine responses, we prevented participants from pasting text into the free-text box. As such, we recruited the following sample size:
\begin{center}
\begin{tabular}{ c c c}
Experiment version  & Condition & Sample size (n) \\
    \hline
Non-autobiographical &    High mood          & \StwoRawAutobiofalseMh{} \\
       &  Low mood &           \StwoRawAutobiofalseMl{} \\
    \hline
Autobiographical    & High mood   &        \StwoRawAutobiotrueMh{} \\
        & Low mood          & \StwoRawAutobiotrueMl{} \\
    \hline
Combined    & High mood   &        \StwoRawMh{}  \\
        & Low mood          & \StwoRawMl{} \\
\hline 
Total &  & \fpeval{\StwoRawMh+\StwoRawMl} \\
\end{tabular}
\end{center}

During data preprocessing, we excluded participants based on the following criteria, specifically those who didn't meet required 50\% word count threshold:
\begin{itemize}
\item had any of PHQ9 missing (either baseline or FU) ($n= \StwoIncompletePhqSub{}$)
\item had any mood rating missing (either baseline or FU) ($n= \StwoIncompleteMoodSub{}$)
\item wrote less than 7 words for any of the recreated diary ($n=\StwoBadSubRec{}$)
\item wrote less than 50 words for the continuation diary ($n=\StwoBadSubInt{}$)
\item wrote less than 15 words in any of the open-ended questions ($n=\StwoBadSubOq{}$)
\item wrote one word repeatedly in any of the open-ended responses ($n=\StwoBadSubOqManual{}$)
\item recalled less than 3 words ($n=\StwoBadRecallSub{}$)
\item wrote full sentences in the recall task ($n=\StwoBadRecallsentenceSub{}$)
\item absolute PHQ9 Q2 change was lesser/greater than 3 standard deviations from the mean for each condition and experiment version ($n=\StwoOutlierPhqSub{}$)
\item absolute mood change was greater than 3 standard deviations from mean for each condition and experiment version ($n=\StwoOutlierMoodSub{}$)
\end{itemize}

This procedure resulted in the final sample size of:
\begin{center}
\begin{tabular}{ c c c c}
Experiment version  & Condition & Sample size (n) & Excluded (n) \\
    \hline
Non-autobiographical &    High mood          & \StwoFinalAutobiofalseMh & \fpeval{\StwoRawAutobiofalseMh - \StwoFinalAutobiofalseMh} \\
       &  Low mood &           \StwoFinalAutobiofalseMl &\fpeval{\StwoRawAutobiofalseMl - \StwoFinalAutobiofalseMl}\\
    \hline
Autobiographical    & High mood   &        \StwoFinalAutobiotrueMh & \fpeval{\StwoRawAutobiotrueMh - \StwoFinalAutobiotrueMh} \\
        & Low mood          & \StwoFinalAutobiotrueMl & \fpeval{\StwoRawAutobiotrueMl - \StwoFinalAutobiotrueMl} \\
    \hline
Combined    & High mood   &        \StwoFinalMh & \fpeval{\StwoRawMh-\StwoFinalMh} \\
        & Low mood          & \StwoFinalMl & \fpeval{\StwoRawMl - \StwoFinalMl}\\
\hline 
Total & & \StwoFinaln & \fpeval{\StwoRawMh-\StwoFinalMh+\StwoRawMl - \StwoFinalMl} \\
\end{tabular}
\end{center}

\subsubsection{End of study questions}
At the end of the study, we asked participants to provide explicit judgements about the following statements on a VAS ("Strongly disagree" - "Strongly Agree"):
\begin{itemize}
    \item Non-Autobiographical
    \begin{enumerate}
        \item I was able to put myself in the other person's shoes after listening to their diary entries.
        \item Recreating and creating diary entries affected my mood.
        \item Recreating and creating diary entries reminded me of similar situations from my own life.
        \item My responses to questions about myself were shaped by what I thought the researcher wanted to hear.
        \item Recreating and creating diary entries did NOT affect my mood.
    \end{enumerate}
    \item Autobiographical
    \begin{enumerate}
        \item I was able to put myself in the other person's shoes after listening to their diary entries.
        \item Listening to and recreating diary entries affected my mood.
        \item The diary entries reminded me of similar situations from my own life.
        \item Creating my own diary entry based on similar life experiences affected my mood.
        \item My responses to questions about myself were shaped by what I thought the researcher wanted to hear.
        \item Creating my own diary entry based on similar life experiences DID NOT affect mood.
    \end{enumerate}
\end{itemize}

\subsection{Study 1 analysis}
\subsubsection{LLMs details}\label{appdx:llm_dets}
We used the Gemma2 2b-it and 9b-it model version \cite{gemma_2024}, as well as Llama-3.1-8B-Instruct and Llama-3.2-3B-Instruct \cite{llama3modelcard}. We also used Mistral-7B-OpenOrca large language model \cite{lian2023mistralorca1}. The model is a fine-tuned version of Mistral-7B-v0.1 model \cite{jiang_mistral_2023}. The original Mistral model was fine-tuned on a rich collection of augmented FLAN data aligns \cite{longpre_flan_2023} based on Orca model \cite{mukherjee_orca_2023}. Essentially the model was fine-tuned on rich signals from GPT-4 model \cite{openai_gpt-4_2023} that include step-by-step thought processes and complex instructions guided by teacher assistance.

\subsubsection{Logit sampling}
We use participants' open-ended question and answer as context for the LLM to responses to the corresponding multiple-choice question on the scale (see prompt in \ref{apdx:item_prompt} We specify that the model has to respond by selecting a character label (A, B, C, D) that corresponds to the questionnaire label response. We do a forward pass through the model to extract the logits at the final token and then select logits corresponding to the (A, B, C, D) tokens. We then covert this restricted logits space into probability using softmax transformation. We then sample 50 responses for each question, each participant.

For generalisation sampling, we slightly change the prompt to include the specific multiple-choice question on the questionnaire scale (SDS, PHQ9, GAD7), while keeping hte open-ended question and answer intact.

\subsubsection{Covariance structure match measures}\label{apdx:cov_match_measures}
In Equation \ref{eq:gen_total_def}, we approximated LLM predicted questionnaire total score for each of the sampled self-report questionnaire, $j$, by summing up scores, $s_{q_{j},c}$, across questionnaire items, $\{q_{j}\}$, for each of the QA-pairs as context, $c$, and then averaging across the QA-pairs.
\begin{equation}
\label{eq:gen_total_def}
	\text{"Totals"} = \hat{T_j} = \frac{1}{C}\sum_{c=1}^{C}\sum_{q_{j}=1}^{Q_j} s_{q_{j},c} 
\end{equation}

Secondly, we quantified the overall correlation-matrix item-level difference by bootstrapping ($B=1000$) the ground-truth ($M^P$) and predicted ($M^L$) matrices. We then calculated the average absolute item-level difference between the two matrices, whose "1 minus value" (complement) (see Equation \ref{eq:gen_item_diff}), indicates how close the values match.

\begin{equation}
\label{eq:gen_item_diff}
	\text{"Item corrs."} = 1-\frac{1}{I\times J}\sum_{i,j}^{I,J}\frac{1}{B}\sum_{b=1}^B\big|M^P_{i,j,b}-M^L_{i,j,b}\big|
\end{equation}

\clearpage
Lastly, we aimed to quantify how well the open-ended verbalisations captured the co-occurrence within different sets of symptoms. For that purpose, we resorted to singular value decomposition (SVD) of the bootstrapped correlation matrices.

\begin{equation}
\label{eq:gen_item_svd_P}
M^P = U^PS^PV^{'P}\\
\end{equation}
\begin{equation}
\label{eq:gen_item_svd_L}
M^L = U^LS^LV^{'L}\\
\end{equation}
Effectively, this decomposition provides a way project two sets of questionnaire scores (using $U$ and $V'$ matrices) into common sub-space and estimates how well they correlate in this new space, as captured by diagonal elements of $S$. By virtue of the SVD, the projection matrices are selected so as to maximise these sub-space component correlations, at the same time capturing the relevant symptom co-occurrences.

This then allows us to gauge how much of the ground-truth sub-component correlations (common sub-space alignment) is reflected in the predicted subspace. We do this by comparing the estimated sub-components correlations, $\hat{S}^L$, given the ground-truth decomposition:
\begin{equation}
\label{eq:gen_item_svd_Shat}
\hat{S}^L = U'^PM^LV^{P}
\end{equation}
to the ground-truth sub-component correlations $S^P$, as
\begin{equation}
	\label{eq:gen_item_svd_Sdelta}
	\Delta S = \hat{S}^L -S^P
\end{equation}

We then define the overall sub-component similarity as the relative sub-component correlation error complement:
\begin{equation}
	\label{eq:gen_item_svd_Shat_err_comp}
	\text{"Singular vals."} = 	1- \frac{|| \Delta S ||_F^2}{|| \hat{S^L} ||_F^2 +|| S^P ||_F^2}
\end{equation}
% We report the bootstrapped "Singular vals." in Fig. \ref{fig:study2_llm_gen_item_qs} Col. IV.

Focusing on the top k sub-components that capture 99\% of the correlations, we can compare two sets of bases ($\{U_{:k}^P,U_{:k}^L\}, \{V_{:k}^P,V_{:k}^L\}$) of the k-dimensional subspace given by decomposition of the ground truth and predicted correlation matrices. We do this by estimating the average "principal angles" between bases within each set by another SVD of the between-bases correlation matrix:
% \clearpage
\begin{equation}
	\label{eq:gen_item_svd_Ubasis_svd}
	U_{:k}^{'P}U_{:k}^L = A^{'U}\Sigma^U B^{U}
\end{equation}

\begin{equation}
	\label{eq:gen_item_svd_Vbasis_svd}
	V_{:k}^{'P}V_{:k}^L = A^{'V}\Sigma^V B^{V}
\end{equation}

This gives the average principle angle for each set as:
\begin{equation}
	\label{eq:gen_item_svd_Ubasis_avg}
	\text{"O. PHQ8 basis"} = \bar{\Sigma}^U = \frac{1}{k}\sum_{i=j}\Sigma^U_{i,j}
\end{equation}

\begin{equation}
	\label{eq:gen_item_svd_Vbasis_avg}
	\text{"[Questionnaire] basis"} = \bar{\Sigma}^V = \frac{1}{k}\sum_{i=j}\Sigma^V_{i,j} 
\end{equation}

Lastly, we can compare the projection matrices formed by the k-dimensional bases of the ground-truth and predicted correlation matrices. A projection matrix provides a description of how much each question contributes the latent-components (the shape of the k-dimensional subspace).

\begin{equation}
	\label{eq:gen_item_svd_Uproj_p}
    P_{U}^P = U_{:k}^P U_{:k}^{'P}
\end{equation}
\begin{equation}
	\label{eq:gen_item_svd_Uproj_p}
    P_{U}^L = U_{:k}^L U_{:k}^{'L}
\end{equation}

\begin{equation}
	\label{eq:gen_item_svd_Vproj_p}
    P_{V}^P = V_{:k}^P V_{:k}^{'P}
\end{equation}
\begin{equation}
	\label{eq:gen_item_svd_Vproj_p}
    P_{V}^L = V_{:k}^L V_{:k}^{'L}
\end{equation}

\clearpage
For each pair of projection matrices, we can then compare the ground-truth and predicted projections using the complement of the normalised Frobenius distance between the two:
\begin{equation}
	\label{eq:gen_item_svd_VP_delta}
    "\text{O. PHQ9 proj.}" = 1 - \Delta P_U = 1 - \frac{|| P_{U}^P-P_{U}^L ||_F}{\sqrt{2k}}
\end{equation}

\begin{equation}
	\label{eq:gen_item_svd_VP_delta}
    "[\text{Questionnaire] proj.}" = 1 - \Delta P_V = 1 - \frac{|| P_{V}^P-P_{V}^L ||_F}{\sqrt{2k}}
\end{equation}
where values closer to 1 indicate stronger alignment between projection matrices.

We rescaled these metrics by $\frac{1}{\sqrt{2k}}$, to constrain the range to 0 and 1, based on the following property of projection matrix distance:
\begin{align}
    ||P^P-P^L||_F^2 = \\
    &= Tr[(P^P-P^L)(P^P-P^L)'] \\
    &\stackrel{\text{symmetry of P}}{=} Tr[(P^P-P^L)(P^P-P^L)]\\
    &=Tr[P{^P}^2]-Tr[2P^P P^L]+Tr[P{^L}^2]\\
    &\stackrel{\text{idempotence}}{=}Tr[P^P]+Tr[P^L]-2Tr[P^P P^L]\\
    &=2k - 2Tr[P^P P^L]
\end{align}
If $P^P$ and $P^L$ are the maximally different (orthogonal), then $Tr[P^P P^L] = 0$, and if there are the same then $Tr[P^P P^L] =k$. Therefore, $||P^P-P^L||_F$ is bound between $0$ and $\sqrt{2k}$.

\subsection{Supervised sparse auto-encoder (sSAE)}\label{apdx:ssae}
We use the ground truth dataset collected in Study 1 to train a supervised sparse auto-encoder (sSAE; \cite{lee_emergence_2025, yun_transformer_2021, le_supervised_2018}) to predict participants' z-scored PHQ-9 scores from the average hidden state representation of the QA-pairs of open-ended prompt.

Specifically, we extracted hidden-states on the token just after the open-ended response for each layer of the model from the middle layer onwards. We averaged the hidden states across eight questions obtaining an average hidden state for each participant, $\vh_i$, pairing this with the true vector of PHQ-9 scores $\vy_i$. 

\subsubsection{Architecture details}
We outline the supervised sparse auto-encoder equations below. We tied together the encoding and decoding matrix weights, the latter being normalised along the hidden dimension and transposed. 

\begin{equation}
    \vh_{cent} = \vh-\vb_{dec}
\end{equation}
\begin{equation}
    \vs_{post} = relu(\vh_{cent}\mW_{enc})
\end{equation}
\begin{equation}
    \vq =\vs_{post}\mW_{proj} + \vb_{proj}
\end{equation}
\begin{equation}
    \vh_{rec} = \vs_{post}\mW_{enc}^T + \vb_{dec}
\end{equation}

Here $\mW_{enc}$ is $d\times (d*f)$, where $d$ is the original hidden state dimensionality from the LLM model $\vh$, while $f$ is the dimension scaling factor.

We defined the following loss items, where $\vy$ is the vector of participant scores ($Q$-dimensional, for PHQ-9 $Q=9$)

Reconstruction loss:
\begin{equation}
    \vl_{rec} = ||(\vh_{rec} - \vh)||^2
\end{equation}
Sparsity loss:
\begin{equation}
    \vl_{sparse} = ||\vs_{post}||_1
\end{equation}
Question prediction loss:
\begin{equation}
    \vl_{qs} = ||(\vq-\vy)||^2
\end{equation}
Questionnaire total severity loss:
\begin{equation}
    \vl_{sev} = \Large|\large|\sum^{Q}{q_j}-\sum^{Q}{y_j}\large|\large|^2
\end{equation}
Total loss
\begin{equation}
    L_{total} = \vl_{rec}+\lambda \vl_{sparse} + \vl_{qs} +  \frac{1}{Q} \vl_{sec}
\end{equation}

\subsubsection{sSAE training}
% We extracted hidden-states on the token just after the open-ended response for each layer of the model from the middle layer onwards. We averaged the hidden states across questions obtaining an average hidden state for each participant, $\vh_i$, we then matched this with the true vector of z-scored PHQ-9 scores $\vy_i$. 

We split the \SsaeNfinal{} participant dataset into training, validation and test set (70\%, 15\%, 15\%). To find the optimal setting of hyper-parameters for the sSAE model \cite{lee_emergence_2025, yun_transformer_2021, le_supervised_2018}, we search the grid for the following hyper-parameters.
\begin{itemize}
    \item Optimiser learning rates: $\alpha \in [0.001, 0.0001]$
    \item SAE sparsity loss coefficient $\lambda \in [0.05, 0.1, 0.2]$
    \item SAE dimension scaling factor $f \in[1,2,4]$
\end{itemize}

We used a batch size of 32. With the training set, we used \cite{kingma_adam_2017} optimiser for each unique setting on the hyper-parameter grid (using back-propagation) to find the best network weights. Then, using the validation set, we found the best hyper-parameter setting for each layer based on the validation loss as well as the correlation between true and predicted questionnaire scores. The final results plotted in the paper are against the held-out test set.

\subsubsection{sSAE structure}\label{apdx:ssae_structure}
To quantify the held-out structure of best-layer sSAE predictions, we predicted each of the PHQ-9 scores given each of the embedding (hidden-state) corresponding to the open-ended symptom verbalisation. We then computed correlations between the predicted and ground-truth score for each question. We computed structure similarity metrics as before (see Methods §\ref{apdx:cov_match_measures})

\subsubsection{sSAE latent score perturbation}\label{apdx:ssae_pert}
In the sSAE perturbation, we aim to achieve a specific change in question  $\Delta\vq_j = [0,..,\delta_j,...,0]$ (here we always chose $\delta_j=1$), such that $\vq_{new,j}=\vq+\Delta \vq_j$. Therefore we seek $\Delta \vs_{post,j}$, where
\begin{equation}
 \vs_{new,j} =\vs_{post}  + \Delta \vs_{post,j}
\end{equation}
such that
\begin{equation}
    \mW_{proj}\vs_{new,j} = \vq_{new,j}
\end{equation}
leading to 
\begin{equation}
    \mW_{proj}\Delta \vs_{post,j} = \Delta \vq_j
\end{equation}
We find $\Delta \vs_{post,j}$ by solving a basis pursuit problem \cite{van_den_berg_probing_2009} using \texttt{spgl1} python package \cite{noauthor_spgl1_nodate}. The goal is to minimise the norm-1 loss, encouraging sparse solutions.

\begin{equation}
    \min || \Delta \vs_{post,j} ||_1 \quad \textit{ subject to } \quad \mW_{proj}\Delta \vs_{post,j} = \Delta \vq_j
\end{equation}

In the main paper, we then plot the perturbed predicted scores on the validation set as a confusion matrix.

\subsubsection{Logits steering using perturbed sSAE latent states}\label{apdx:logit_pert}
Here we want to perturb the hidden states to change the severity of the multiple-choice responses given the open-ended response. For each question $j$, we extract the hidden states at each layer for the token just after the open-ended response and at the last token just before the response label is selected and average: $\vh_{avg,j}$. We then pass the $\vh_{avg,j}$ through the sSAE to perturb the latent state:$\vs_{post,avg,j} + \Delta \vs_{post_j}$, we then reconstruct this back to the original hidden-state space $\vh_{avg,pert,j}$.

At the last token of the input sequence, we alter the original hidden state $h_{last}$ sequentially, one layer at a time in the forward pass as follows. We first project the reconstructed perturbed sSAE state onto the original hidden state: 
% \begin{equation}
%     \vp = \vh_{last} (\frac{\vh_{avg,pert,j}}{||\vh_{avg,pert,j}||} \cdot \vh_{last}  ) \frac{1}{||\vh_{last}||}
% \end{equation}
\begin{equation}
    \vp = \vh_{last} (\vh_{avg,pert,j} \cdot \vh_{last}  ) \frac{1}{||\vh_{last}||^2}
\end{equation}

and find the difference between the projection and the desired direction
\begin{equation}
    \vv = \vh_{avg,pert,j} - \vp
\end{equation}
We then add this difference to the original hidden state to move it in the desired direction with strength $\gamma$ (if the dot product was negative, we flip the direction of the hidden state first):
\begin{equation}
    \vh_{last,pert} = sign(\vh_{avg,pert,j} \cdot \vh_{last})\vh_{last} + \gamma \vv
\end{equation}

Using the validation test we considered $\gamma \in [-1.5,-1,-0.5,-0.25,0.25,0.5,1,1.5]$. We looked for strongest Cohen-d effect sizes in the expected scores $\Delta y_{i,j} = \mathrm{E}[y_{i,j,pert} -y_{i,j}]$ and found that $\gamma \in [{1.5, -0.25}]$ provides the most robust changes. We then resampled the perturbed responses using the held-out test set, using these strength settings, and plotted the results in the main paper.

Finally, for each participant, for each question, we calculate the expected score based on the perturbed logits and compare to expected score of unperturbed logits. For each question we then calculate Cohen's effect size of deviation from the unperturbed scores across the test set and compute one-sided t-test.

\subsection{Study 2 analysis}
\subsubsection{Text similarity}
To quantify the similarity metrics for participants' recreated diaries and diary continuations, we used \texttt{SentenceTransformer "all-MiniLM-L6-v2} \cite{reimers-2019-sentence-bert} to represents the texts as an embedding vector. We then computed cosine similarity between vectors for each pair of: 1) original diary transcript and recreated diary; 2) diary continuation and recreated diaries (averaged across diaries).

\subsubsection{Latent sSAE measure}
For each participant's continuation diary entry, we extracted Gemma2\_9B's hidden state at each layer (from 22 to 42) on the last token of the continuation diary entry, which we then normalised to length 1. We then applied our sSAE model to obtain layer-wise sSAE PHQ9 scores for each text. The average sSAE score across layers for each participant was then used for the analyses.

\subsubsection{Word recall analysis}
We used SiEBERT - English-Language Sentiment Classification model \cite{hartmann2023} to calculate average sentiment score across all recalled words. We calculated average baseline recall sentiment across two baseline instances of recall, which we then compared to average FU recall sentiment.

\backmatter
% \bmhead{Acknowledgements}

\section*{Declarations}
JO was supported by the International Max Planck Research School on Computational Methods in Psychiatry and Ageing Research (IMPRS COMP2PSYCH) fellowship (577749/D-CON/186534). 
Part of this work was supported by a Wellcome Trust grant to QJMH (221826/Z/20/Z). QJMH was employed by University College London during this work.
QJMH has obtained fees and options for consultancies for Aya Technologies and Alto Neuroscience.
QJMH has received research grant funding from Carigest S.A., German Research Foundation, Koa Health,
NIHR, Swiss National Science Foundation, Wellcome Trust. QJMH acknowledges support by the NIHR UCLH BRC and NIHR MH-TRC MHM.
JO declares no competing interests.\\

Code is available here: \url{https://github.com/onyskj/internal_narratives}.\\

Data is available here: \url{https://osf.io/bs2df}. Note that open-ended free text responses, as well as corresponding LLM hidden state files are not included due to anonymity concern. We are happy to set up Data Sharing Agreement to share these instead. Please contact the corresponding author.

\clearpage
% \bibliography{sn-bibliography}% common bib file
\bibliography{clean_references.bib}% common bib file

\noindent
\bigskip

\clearpage
% \newpage

\begin{appendices}
\bmhead{Supplementary information - Metareasoning constraints couple narratives, affect and cognition}
\section{Demographics}

\subsection{Severity histograms}

\begin{center}
\begin{figure}[!hbt]
    % \centering
    \centerfloat
    % \makebox[\textwidth][c]{\includegraphics[width=1.2\textwidth]{figs//fig_supp/test1_qs_structure_totals.pdf}}%
    \includegraphics{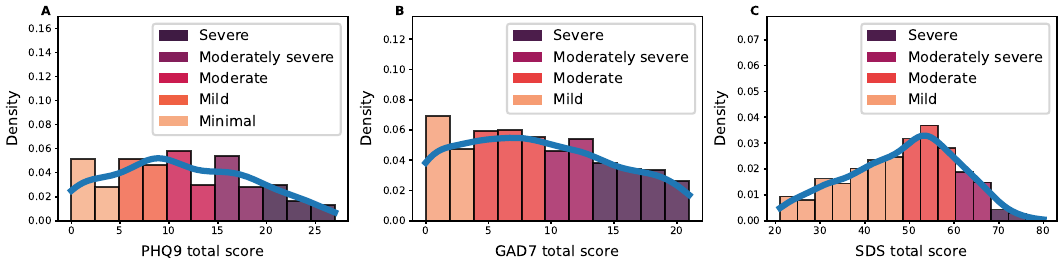}
    \caption{Study 1 total scores histograms for \textbf{A}: PHQ-9 (depression), \textbf{B}: GAD-7 (anxiety); \textbf{C}: SDS (depression). Severity was classified based on the following total score ($t$) cut-offs. PHQ-9: $t<=4$ (Minimal); $4<t<=9$ (Mild); $9<t<=14$ (Moderate); $14<t<=19$ (Moderately severe); $19<t<=27$ (Severe). GAD-7: $t<=4$ (Mild); $4<t<=9$ (Moderate); $9<t<=14$ (Moderate severe); $14<t<=21$ (Severe). SDS: $t<=49$ (Mild); $49<t<59=$ (Moderate); $59<t<=69$ (Moderate severe); $69<t<=80$ (Severe)}
    \label{fig:apdx_s1_totals}
\end{figure}
\end{center}

\begin{figure}[!hbt]
    \centering
    \includegraphics{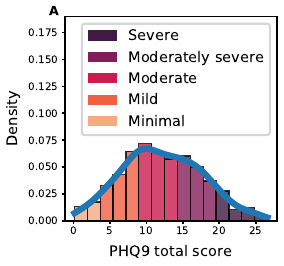}
    \caption{
    Study 2 total scores histograms for \textbf{A}: PHQ-9 (depression). Severity was classified based on the following total score ($t$) cut-offs. PHQ-9: $t<=4$ (Minimal); $4<t<=9$ (Mild); $9<t<=14$ (Moderate); $14<t<=19$ (Moderately severe); $19<t<=27$ (Severe).
    }
    \label{fig:apdx_s2_totals}
\end{figure}

\clearpage
\subsection{Demographics table}
\begin{center}
\renewcommand{\arraystretch}{1.2}
\begin{table}[!hbt]
		\begin{tabular}{|l|c|c|c|c|c|}
			\hline
			\multicolumn{1}{|c|}{Demographics} & \multicolumn{1}{c|}{Study 1} & \multicolumn{2}{c|}{Study 2 Non-autobiographical} & \multicolumn{2}{c|}{Study 2 Autobiographical} \\
			% \hline
			  \multicolumn{1}{|c|}{} & \multicolumn{1}{c|}{} & High mood & Low mood & High mood & Low mood \\
			\hline
			\hline
            \TableDemographics
			\hline
		\end{tabular}
\caption{Study 1 and 2 demographics summary.}
\label{apdx:demo_table}
\end{table}
\end{center}

\section{Study 1}
\subsection{Open-ended questions}\label{apdx:open_qs}
\begin{enumerate}
    \item Could you share any activities or events from the past two weeks that made you feel bothered because of a lack of interest or pleasure in doing them?
    \item For the past two weeks, have you been bothered by your mood and how you felt generally? Were there any situations when you felt down, depressed, or hopeless?
    \item Can you provide examples of how your sleep has been in the past two weeks? Have you been bothered by challenges with falling asleep, staying asleep, or even sleeping too much?
    \item Have you been bothered by your energy levels over the past two weeks? Can you recall situations when it comes to feeling tired/lively or low/high on energy?
    \item In the past two weeks, have you been bothered about your appetite? Can you describe your typical attitude towards food - maybe you have noticed something unusual, like changes in how much you're eating or not eating?
    \item In the past two weeks, have you been bothered by feelings about yourself? In what situations did you feel proud or like a failure? Did you feel you met your own and your family's expectations, or let them down?
    \item In the past two weeks, have you been bothered by your ability to concentrate and focus? Please describe how it felt to do things that require you to concentrate for a while, like working, reading, or watching movies?
    \item Can you describe situations over the past two weeks when you were bothered by feeling slower than usual in terms of thinking, speaking, or just acting - or situations where you felt fidgety and restless?
\end{enumerate}

\subsection{Multiple-choice questions}\label{apdx:closed_qs}
\clearpage

\subsubsection{PHQ-9 items}
Participants were instructed to consider "Over the last 2 weeks, how often have you been bothered by any of the following problems?". These are answered on a scale - Not at all (0), Several days (1), More than half the days (2), Nearly every day (3) \cite{kroenke_phq-9_2001}.
\begin{enumerate}
    \item Little interest or pleasure in doing things.
    \item Feeling down, depressed, or hopeless.
    \item Trouble falling or staying asleep, or sleeping too much.
    \item Feeling tired or having little energy
    \item Poor appetite or overeating.
    \item Feeling bad about yourself - or that you are a failure or have let yourself or your family down.
    \item Trouble concentrating on things, such as reading the newspaper or watching television.
    \item Moving or speaking so slowly that other people could have noticed? Or the opposite - being so fidgety or restless that you have been moving around a lot more than usual.
    \item Thoughts that you would be better off dead or of hurting yourself in some way.
\end{enumerate}

\subsubsection{SDS items} \label{apdx:sds_questions}
These are answered on a scale - A little of the time (1), Some of the time (2), Good part of the time (3), Most of time (4). Questions 2, 5, 6, 11, 12, 14, 16, 17, 18 and 20 are reverse scored \cite{zung_self-rating_1965}.
\begin{enumerate}

\item I feel down-hearted and blue
\item Morning is when I feel the best
\item I have crying spells or feel like it
\item I have trouble sleeping at night
\item I eat as much as I used to
\item I still enjoy sex
\item I notice that I am losing weight
\item I have trouble with constipation
\item My heart beats faster than usual
\item I get tired for no reason
\item My mind is as clear as it used to be
\item I find it easy to do the things I used to
\item I am restless and can’t keep still
\item I feel hopeful about the future
\item I am more irritable than usual
\item I find it easy to make decisions
\item I feel that I am useful and needed
\item My life is pretty full
\item I feel that others would be better off if I were dead
\item I still enjoy the things I used to do
\end{enumerate}

\subsubsection{GAD-7 items} \label{apdx:sds_questions}
Participants were instructed to consider "Over the last 2 weeks, how often have you been bothered by any of the following problems?". These are answered on a scale - Not at all (0), Several days (1), More than half the days (2), Nearly every day (3) \cite{spitzer_brief_2006}.
\begin{enumerate}
\item Feeling nervous, anxious or on edge
\item Not being able to stop or control worrying
\item Worrying too much about different things
\item Trouble relaxing
\item Being so restless that it is hard to sit still
\item Becoming easily annoyed or irritated
\item Feeling afraid as if something awful might happen
\end{enumerate}

\subsection{Example item-level LLM prompt}\label{apdx:item_prompt}
We used a system prompt "You are a human participant in a study." Below is the prompt we used to sample responses from the logits of each response label character (A, B, C, D).
\begin{lstlisting}
<start_of_turn>user
Please answer the following question in detail.

Question:
For the past two weeks, have you been bothered by your mood and how you felt generally? Were there any situations when you felt down, depressed, or hopeless?

Answer:<end_of_turn>
<start_of_turn>model
 *** Participant's response redacted*** <end_of_turn>
<start_of_turn>user
Please answer the following question.

Over the last 2 weeks, how often have you been bothered the following problem?

Problem: Feeling down, depressed, or hopeless.

Please answer by responding only with one letter (A, B, C, D) from the scale below:
A - Not at all
B - Several days
C - More than half the days
D - Nearly every day

Answer:<end_of_turn>
<start_of_turn>model
\end{lstlisting}

\subsection{LLM sampling}

\subsubsection{Logit sampling}
We use participants' open-ended question and answer as context for the LLM to responses to the corresponding multiple-choice question on the scale (see prompt in \ref{apdx:item_prompt} We specify that the model has to respond by selecting a character label (A, B, C, D) that corresponds to the questionnaire label response. We do a forward pass through the model to extract the logits at the final token and then select logits corresponding to the (A, B, C, D) tokens. We then covert this restricted logits space into probability using softmax transformation. We then sample 50 responses for each question, each participant.

\clearpage
\subsubsection{Item level sampling - remaining models models results}\label{adpx:itemlogit_extra}
\paragraph{Gemma2\_2B}
\begin{figure}[!hbt]
    \centering
    \centerfloat
    \includegraphics{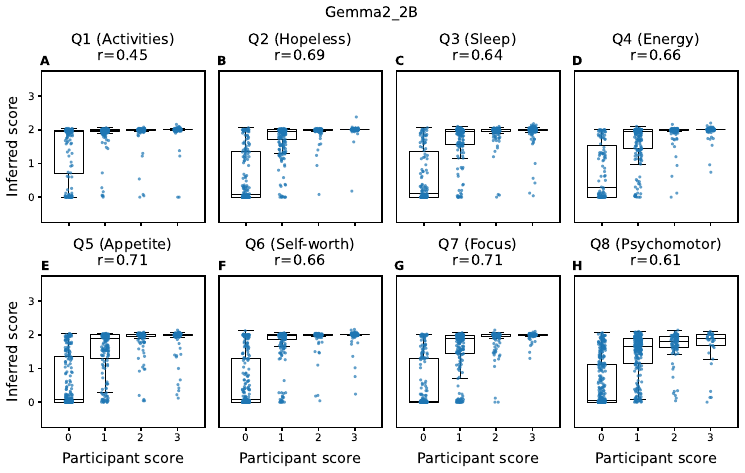}
    \caption{\textbf{PHQ-8 item-level Gemma2\_2B predictions} \textbf{A-H}: Box and scatter plots, with correlations of participant's item-level scores vs Gemma2\_2B scores for the PHQ-8 items given the corresponding open-ended QA-pair}
\end{figure}

\paragraph{Llama-3.1\_8B}
\begin{figure}[!hbt]
    \centering
    \centerfloat
    \includegraphics{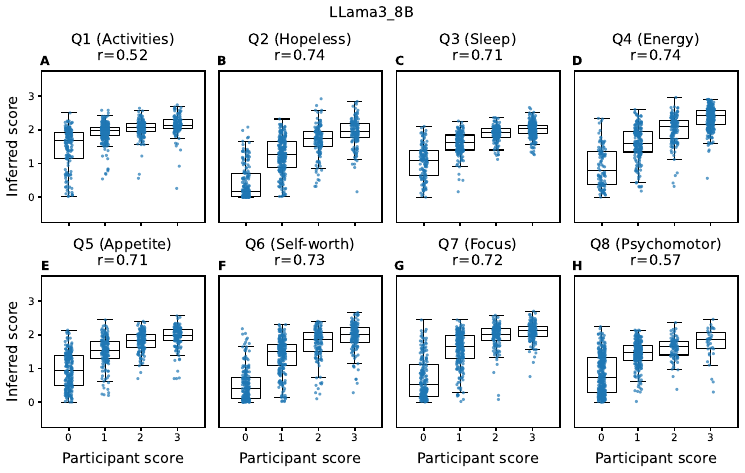}
    \caption{\textbf{PHQ-8 item-level Llama-3.1\_8B predictions} \textbf{A-H}: Box and scatter plots, with correlations of participant's item-level scores vs Llama-3.1\_8B scores for the PHQ-8 items given the corresponding open-ended QA-pair}
\end{figure}

\clearpage
\paragraph{Llama-3.2\_3B}
\begin{figure}[!hbt]
    \centering
    \centerfloat
    \includegraphics{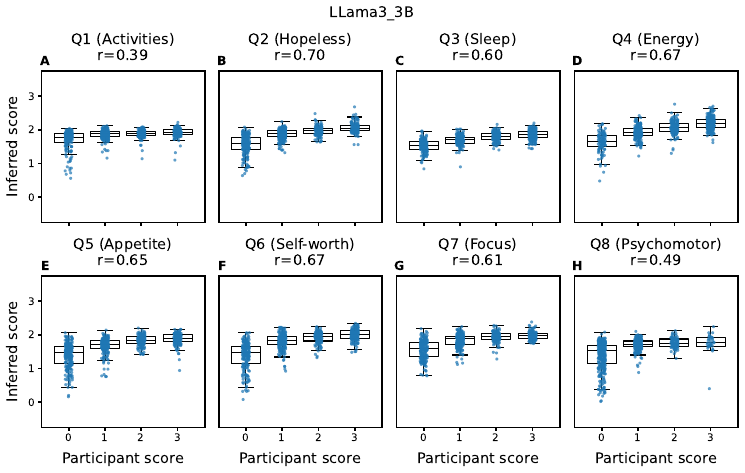}
    \caption{\textbf{PHQ-8 item-level Llama-3.2\_3B predictions} \textbf{A-H}: Box and scatter plots, with correlations of participant's item-level scores vs Llama-3.2\_3B scores for the PHQ-8 items given the corresponding open-ended QA-pair}
\end{figure}

\paragraph{Mistral-7B-OpenOrca}
\begin{figure}[!hbt]
    \centering
    \centerfloat
    \includegraphics{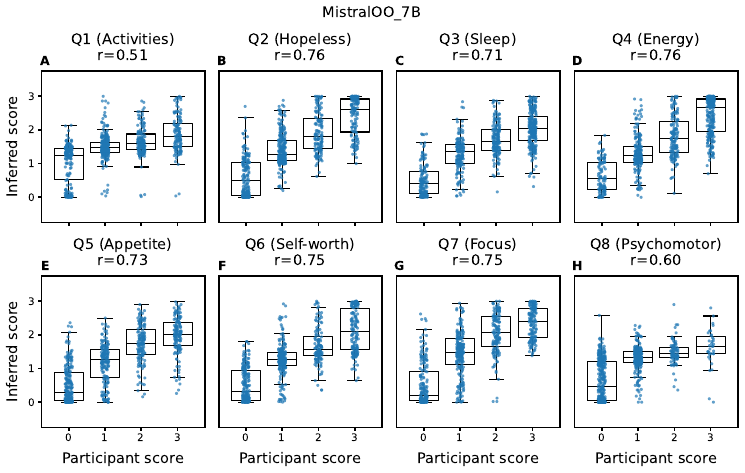}
    \caption{\textbf{PHQ-8 item-level Mistral-7B-OpenOrca predictions} \textbf{A-H}: Box and scatter plots, with correlations of participant's item-level scores vs Mistral-7B-OpenOrca scores for the PHQ-8 items given the corresponding open-ended QA-pair}
\end{figure}

\clearpage

\subsubsection{Item level sampling biases}\label{adpx:itemlogit_extra_bias}
\paragraph{Gemma2\_9B}
\begin{figure}[!hbt]
    \centering
    \centerfloat
    \includegraphics{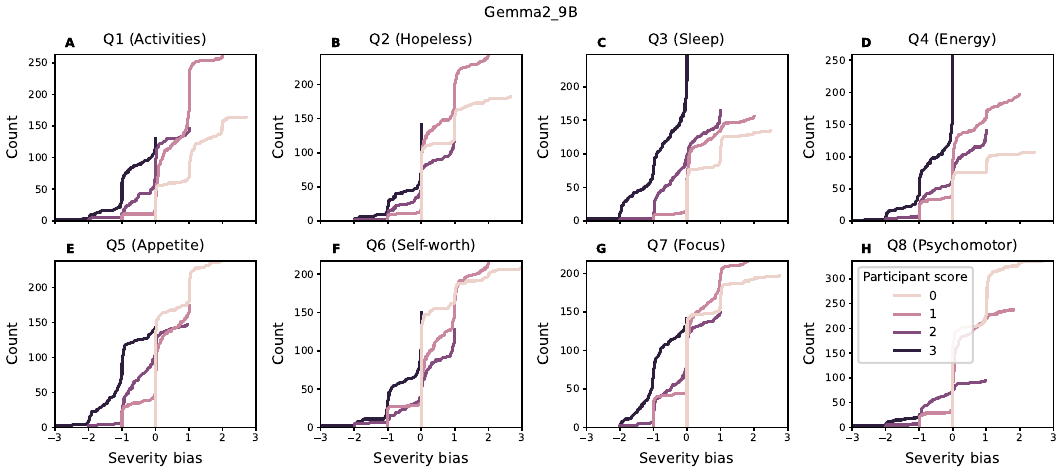}
    \caption{\textbf{PHQ-8 item-level Gemma2\_9B prediction biases}. \textbf{A-H}: Cumulative count plot of severity bias for Gemma2\_9B for each ground-truth score for each question.}
\end{figure}

\paragraph{Gemma2\_2B}
\begin{figure}[!hbt]
    \centering
    \centerfloat
    \includegraphics{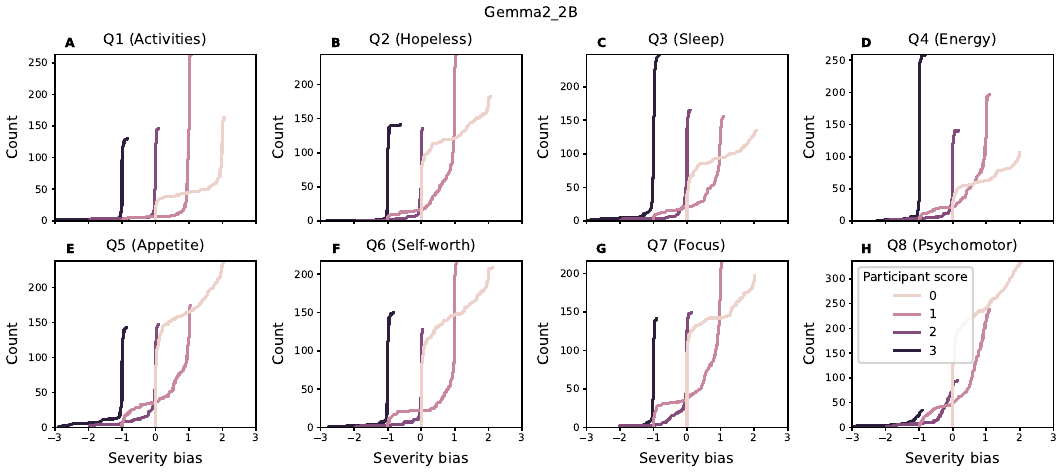}
    \caption{\textbf{PHQ-8 item-level Gemma2\_2B prediction biases}. \textbf{A-H}: Cumulative count plot of severity bias for Gemma2\_2B for each ground-truth score for each question.}
\end{figure}

\clearpage

\paragraph{Llama-3.1\_8B}
\begin{figure}[!hbt]
    \centering
    \centerfloat
    \includegraphics{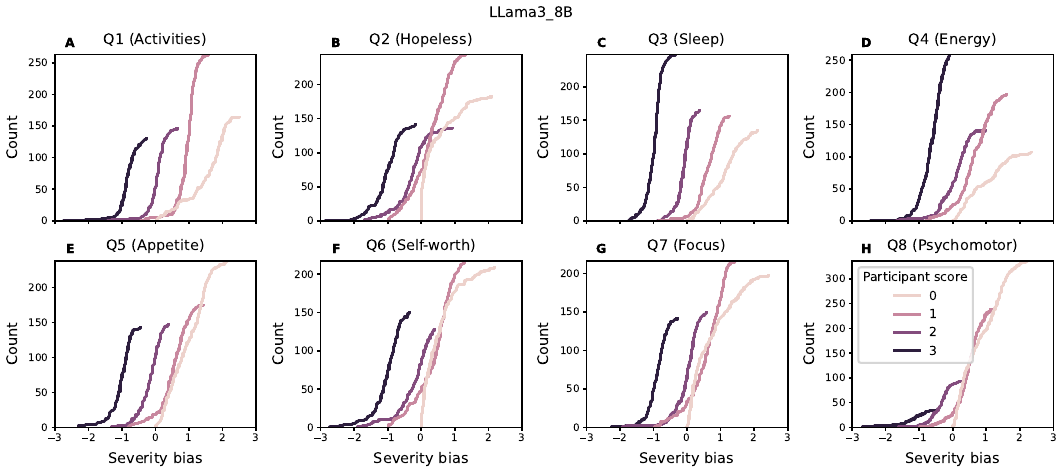}
    \caption{\textbf{PHQ-8 item-level Llama-3.1\_8B prediction biases}. \textbf{A-H}: Cumulative count plot of severity bias for Llama-3.1\_8B for each ground-truth score for each question.}
\end{figure}

\paragraph{Llama-3.2\_3B}
\begin{figure}[!hbt]
    \centering
    \centerfloat
    \includegraphics{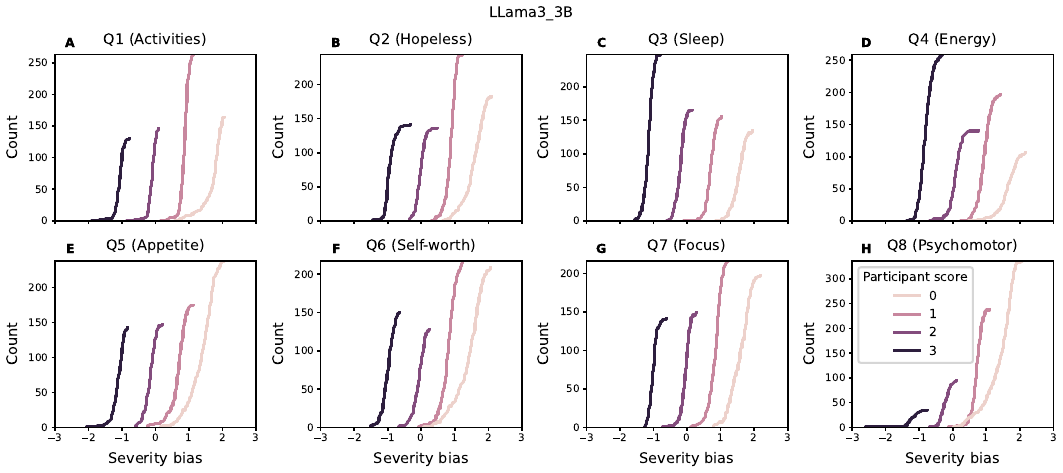}
    \caption{\textbf{PHQ-8 item-level Llama-3.2\_8B prediction biases}. \textbf{A-H}: Cumulative count plot of severity bias for Llama-3.2\_3B for each ground-truth score for each question.}
\end{figure}

\clearpage
\paragraph{Mistral-7B-OpenOrca}
\begin{figure}[!hbt]
    \centering
    \centerfloat
    \includegraphics{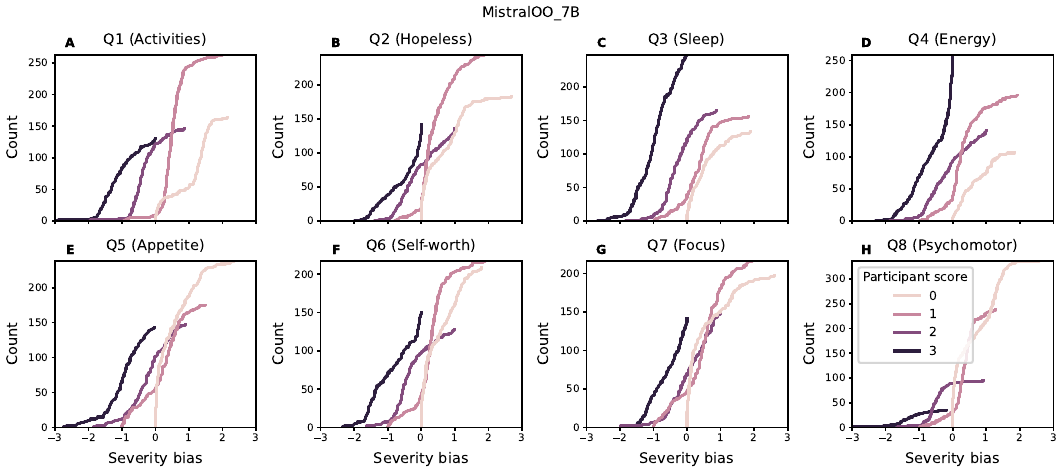}
    \caption{\textbf{PHQ-8 item-level Mistral-7B-OpenOrca prediction biases}. \textbf{A-H}: Cumulative count plot of severity bias for Mistral-7B-OpenOrca for each ground-truth score for each question.}
\end{figure}

\clearpage
\subsubsection{Generalisation sampling - remaining models models results}\label{adpx:gen_logit_extra}

\paragraph{Gemma2\_2B}
\begin{figure}[!hbt]
    \centering
    \centerfloat
    \includegraphics{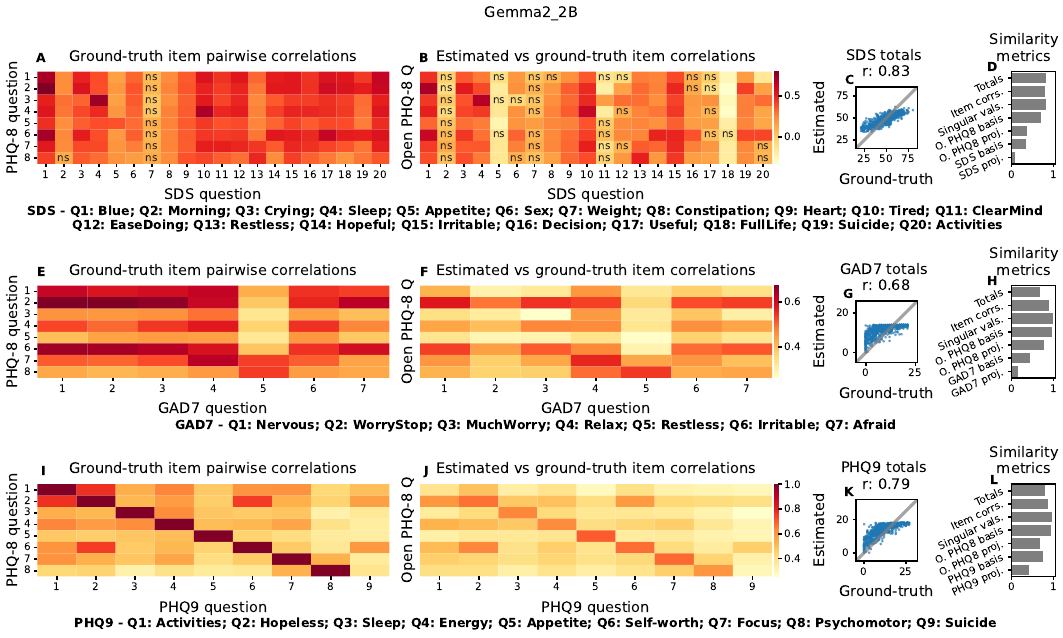}
       \caption{\textbf{Comparison between participant and Gemma2\_2B questionnaires' latent structure} 
       \textit{Column I}: Participants’ ground-truth pairwise correlations between PHQ-8 item scores and SDS, GAD7 and PHQ9 scores (ns - non-significant correlations). \textit{Column II}: Correlations between true SDS, GAD7, PHQ9 item scores (x-axis) and Gemma2\_9B recovered scores for that question given each of the open-ended PHQ-8 QA-pairs (y-axis). \textit{Column III}:  Participants' total scores for the SDS, GAD7 and PHQ9 questionnaire against total score estimate (Pearson correlation). \textit{Column IV}: Similarity metrics quantifying how close the LLM-recovered questionnaire structure matches the participants structure for SDS, GAD7 and PHQ9 (values closer to 1 indicate a more similar response structure) - total scores correlations, average 1 - item-level correlations difference, singular value decomposition bases and projection measures.}
\end{figure}

\newpage
\paragraph{Llama-3.1\_8B}
\begin{figure}[!hbt]
    \centering
    \centerfloat
    \includegraphics{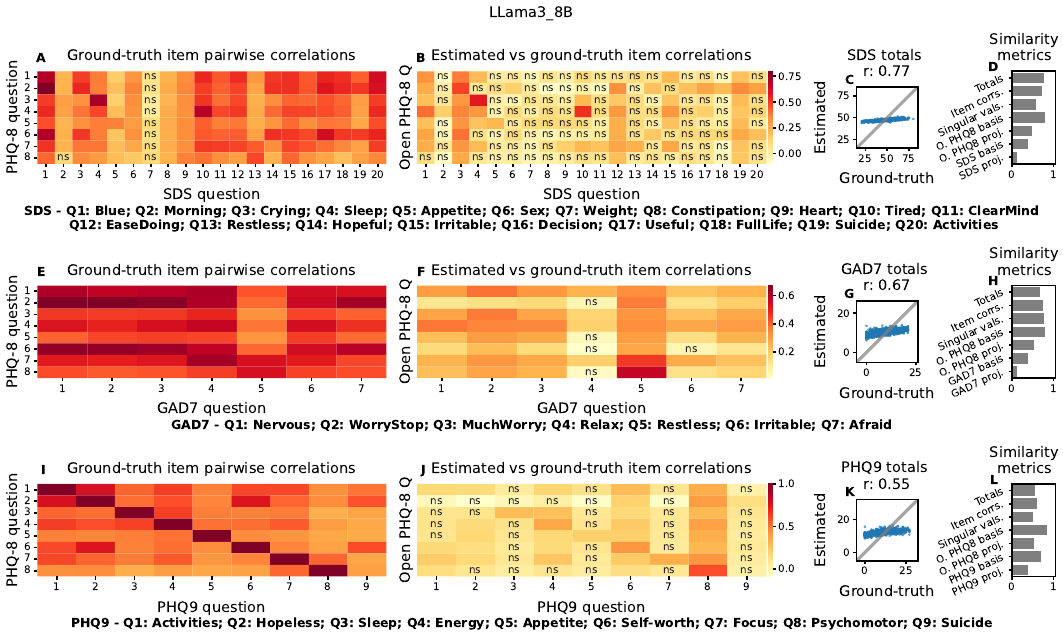}
       \caption{\textbf{Comparison between participant and Llama-3.1\_8B questionnaires' latent structure} 
       \textit{Column I}: Participants’ ground-truth pairwise correlations between PHQ-8 item scores and SDS, GAD7 and PHQ9 scores (ns - non-significant correlations). \textit{Column II}: Correlations between true SDS, GAD7, PHQ9 item scores (x-axis) and Llama-3.1\_8B recovered scores for that question given each of the open-ended PHQ-8 QA-pairs (y-axis). \textit{Column III}:  Participants' total scores for the SDS, GAD7 and PHQ9 questionnaire against total score estimate (Pearson correlation). \textit{Column IV}: Similarity metrics quantifying how close the LLM-recovered questionnaire structure matches the participants structure for SDS, GAD7 and PHQ9 (values closer to 1 indicate a more similar response structure) - total scores correlations, average 1 - item-level correlations difference, singular value decomposition bases and projection measures.}
\end{figure}

\newpage
\paragraph{Llama-3.2\_3B}
\begin{figure}[!hbt]
    \centering
    \centerfloat
    \includegraphics{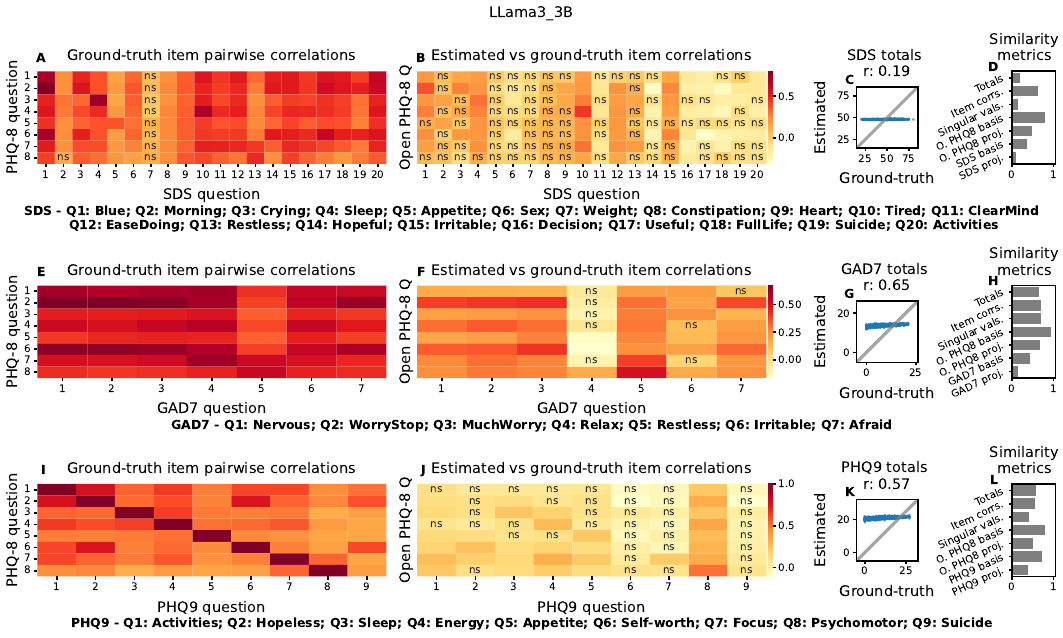}
       \caption{\textbf{Comparison between participant and Llama-3.2\_3B questionnaires' latent structure} \textit{Column I}: Participants’ ground-truth pairwise correlations between PHQ-8 item scores and SDS, GAD7 and PHQ9 scores (ns - non-significant correlations). \textit{Column II}: Correlations between true SDS, GAD7, PHQ9 item scores (x-axis) and Llama-3.2\_3B recovered scores for that question given each of the open-ended PHQ-8 QA-pairs (y-axis). \textit{Column III}:  Participants' total scores for the SDS, GAD7 and PHQ9 questionnaire against total score estimate (Pearson correlation). \textit{Column IV}: Similarity metrics quantifying how close the LLM-recovered questionnaire structure matches the participants structure for SDS, GAD7 and PHQ9 (values closer to 1 indicate a more similar response structure) - total scores correlations, average 1 - item-level correlations difference, singular value decomposition bases and projection measures.}
\end{figure}

\newpage
\paragraph{Mistral-7B-OpenOrca}
\begin{figure}[!hbt]
    \centering
    \centerfloat
    \includegraphics{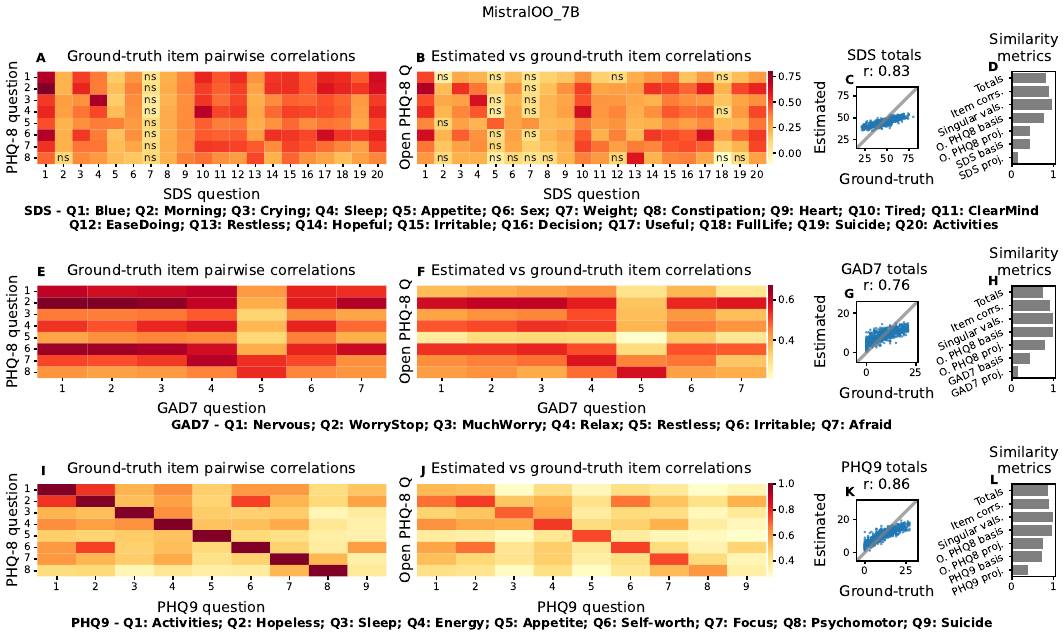}
       \caption{\textbf{Comparison between participant and Mistral-7B-OpenOrca questionnaires' latent structure}  \textit{Column I}: Participants’ ground-truth pairwise correlations between PHQ-8 item scores and SDS, GAD7 and PHQ9 scores (ns - non-significant correlations). \textit{Column II}: Correlations between true SDS, GAD7, PHQ9 item scores (x-axis) and Mistral-7B-OpenOrca recovered scores for that question given each of the open-ended PHQ-8 QA-pairs (y-axis). \textit{Column III}:  Participants' total scores for the SDS, GAD7 and PHQ9 questionnaire against total score estimate (Pearson correlation). \textit{Column IV}: Similarity metrics quantifying how close the LLM-recovered questionnaire structure matches the participants structure for SDS, GAD7 and PHQ9 (values closer to 1 indicate a more similar response structure) - total scores correlations, average 1 - item-level correlations difference, singular value decomposition bases and projection measures.}
\end{figure}

\begin{figure}[!hbt]
    \centering
    \centerfloat
            \includegraphics{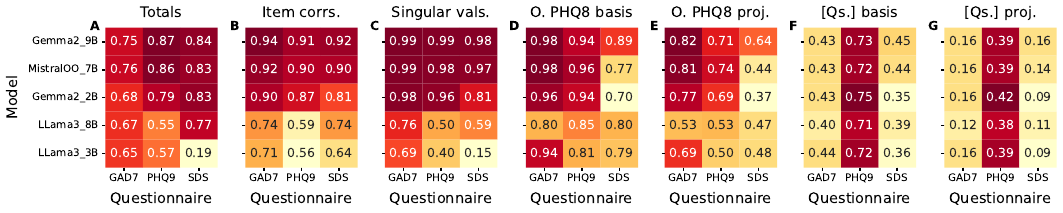}
        \caption{Across model and questionnaire similarity metrics quantifying how close the LLM-recovered questionnaire structure matches the participants structure (values closer to 1 indicate a more similar response structure) - \textbf{A}: total scores correlations, \textbf{B}: average 1 - item-level correlations difference, \textbf{C-G} singular value decomposition bases and projection measures.}
        \label{apdx:study1_cov_metrics_all}
\end{figure}

\clearpage
\section{Supervised sparse auto-encoder study}

\begin{figure}[!hbt]
    \centering
    \centerfloat
    \includegraphics[width=1\linewidth]{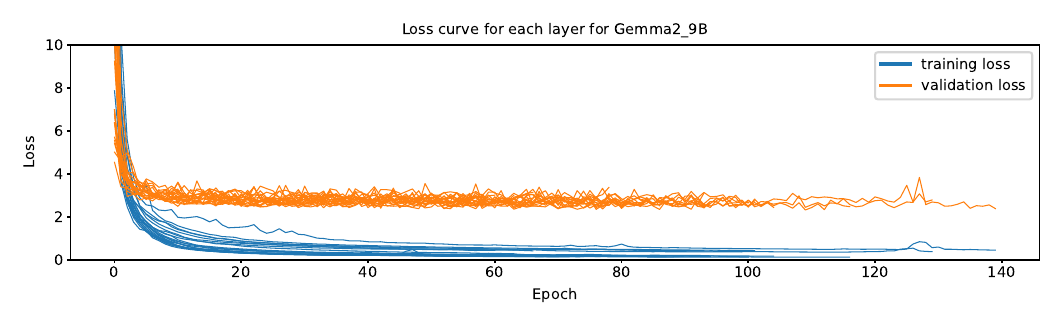}
    \caption{
    Plot of training and validation loss for the best hyper-parameter setting sSAE based on Gemma2-9B hidden state to predict PHQ-9 z-scores. Each line corresponds to layer-specific SAE.
    }
    \label{fig:placeholder}
\end{figure}

\begin{figure}[!hbt]
    \centering
    \centerfloat
    \includegraphics[width=1\linewidth]{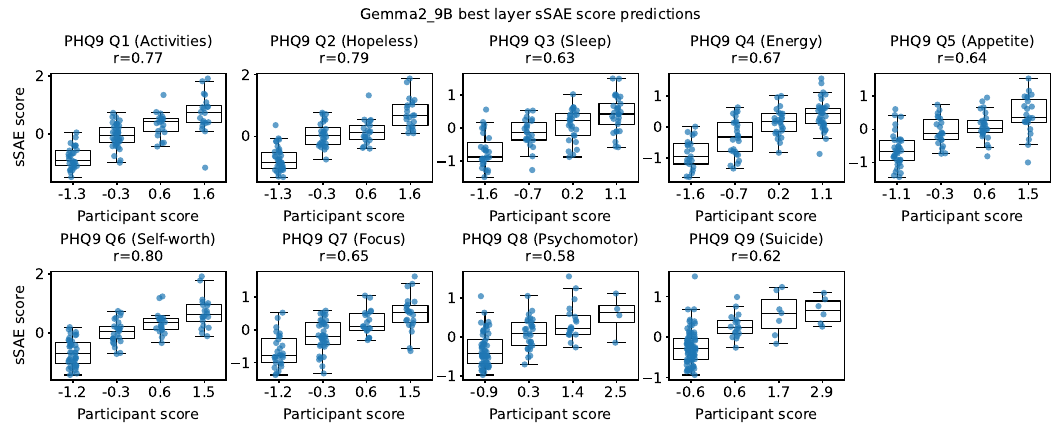}
    \caption{
    \textbf{A-I}: Box and scatter plots, with correlations (p $<0.001$) of subject's item-level z-scores vs the best performing sSAE model (layer 41, Gemma2-9B) predicted scores for PHQ-9 items, given the average hidden state across all open-ended responses for each participant.}
    \label{fig:placeholder}
\end{figure}
\clearpage

\begin{figure}[!hbt]
    \centering
    \centerfloat
    \includegraphics[width=1\linewidth]{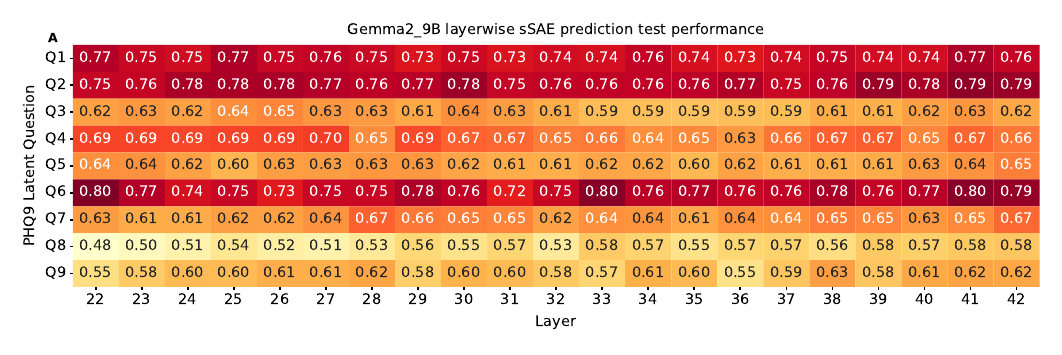}
    \caption{
    \textbf{A}: Correlation between participants' item-level scores and sSAE predicted scores for each layer and each question.}
    \label{fig:placeholder}
\end{figure}

\begin{figure}[!hbt]
        \centering
    \centerfloat
            \includegraphics[width=1\textwidth]{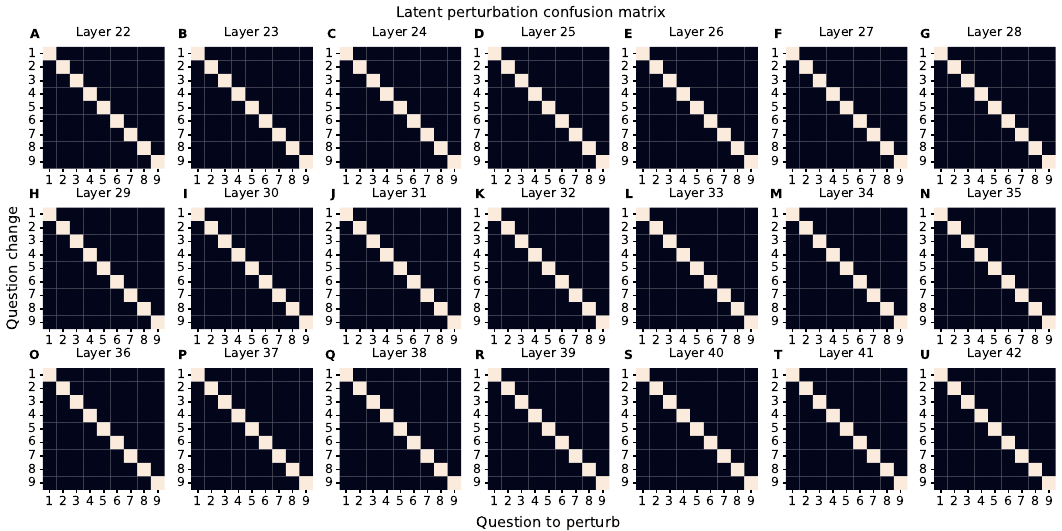}
        \caption{Confusion matrices for latent sSAE PHQ9 perturbation analysis across all layers. For each x-axis question, we identify a perturbation to the sparse representation that results in a specific change to the sSAE latent PHQ9 score (y-axis).}
        \label{fig:ssae_apdx_pert}
\end{figure}

\clearpage

\begin{figure}[!hbt]
        \centering
    \centerfloat
            \includegraphics[width=1\linewidth]{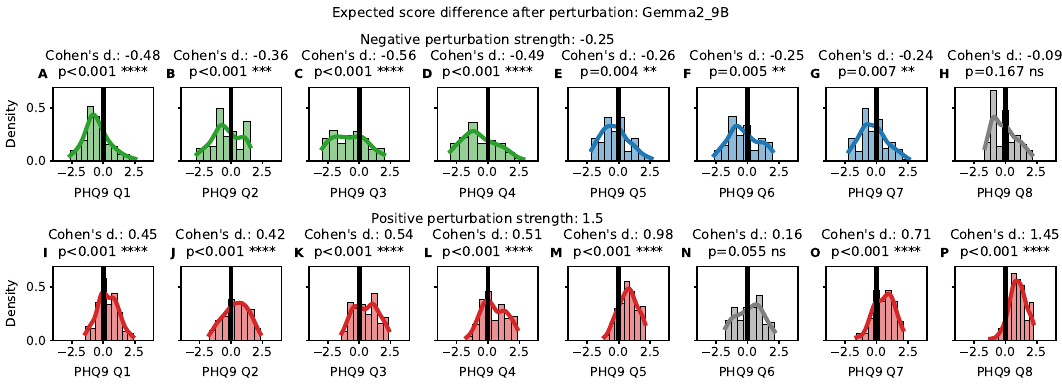}
        \caption{
            The effect of hidden states sSAE perturbation on sampled item-level PHQ-9 responses, given participants' corresponding open-ended QA-pair. We plot histograms of differences between expected steered sampled scores and original scores for \textbf{A-H}: positive direction (less symptom severity) and \textbf{I-P} negative direction (more symptom severity). For each, we report the Cohen-d effect size and the p-value of one-sided t-test. We use blue for small effects $|d|<0.3$, green for positive direction effects $d<-0.3$ and red for negative direction effects $d>0.3$. 
        }
        \label{fig:}
\end{figure}

% \clearpage

\section{Study 2}\label{apdx:s3_section}
\subsection{Measures}\label{apdx:s3_proc}
\subsubsection{Open-ended questions}\label{apdx:open_qs_s2}
At baseline and after mood induction, we asked participant the following questions that they had to answer in an open-ended manner, similarly as in Study 1. They had to write a minimum of 30 words under 100 seconds.
\begin{enumerate}
    \item For the past two weeks, have you been bothered by your mood and how you felt generally? Were there any situations when you felt down, depressed, or hopeless?
    \item Have you been bothered by your energy levels over the past two weeks? Can you recall situations when it comes to feeling tired/lively or low/high on energy?
\end{enumerate}

\subsubsection{Momentary mood}
\begin{figure}[!hbt]
    \centering
    \centerfloat
    \includegraphics[width=1\linewidth]{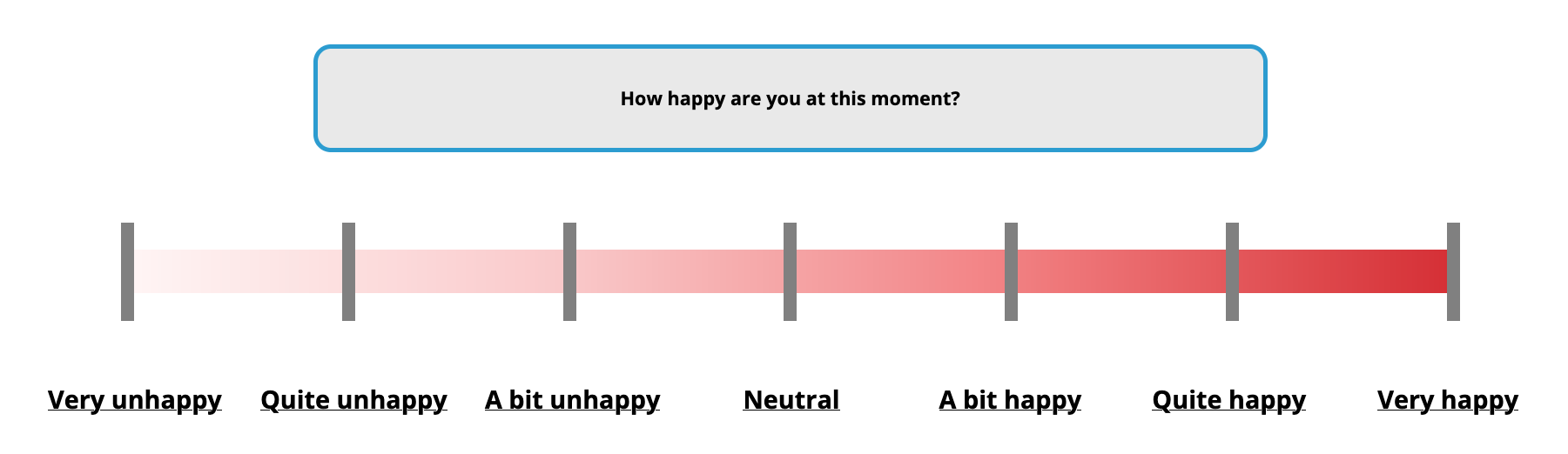}
    \caption{Screenshot of the mood rating VAS scale used in Study 3.}
    \label{fig:apdx:vas_moood}
\end{figure}

Before and after mood induction, participants were asked to rate their momentary mood by answering a question "How happy are you at this moment?" on a continuos visual analogue scale ranging labels: Very unhappy, Quite unhappy, A bit unhappy, Neutral, A bit happy, Quite happy, Very happy. See Fig \ref{fig:apdx:vas_moood}.

\subsubsection{PHQ-9 VAS}
\begin{figure}[!hbt]
    \centering
    \centerfloat
    \includegraphics[width=1\linewidth]{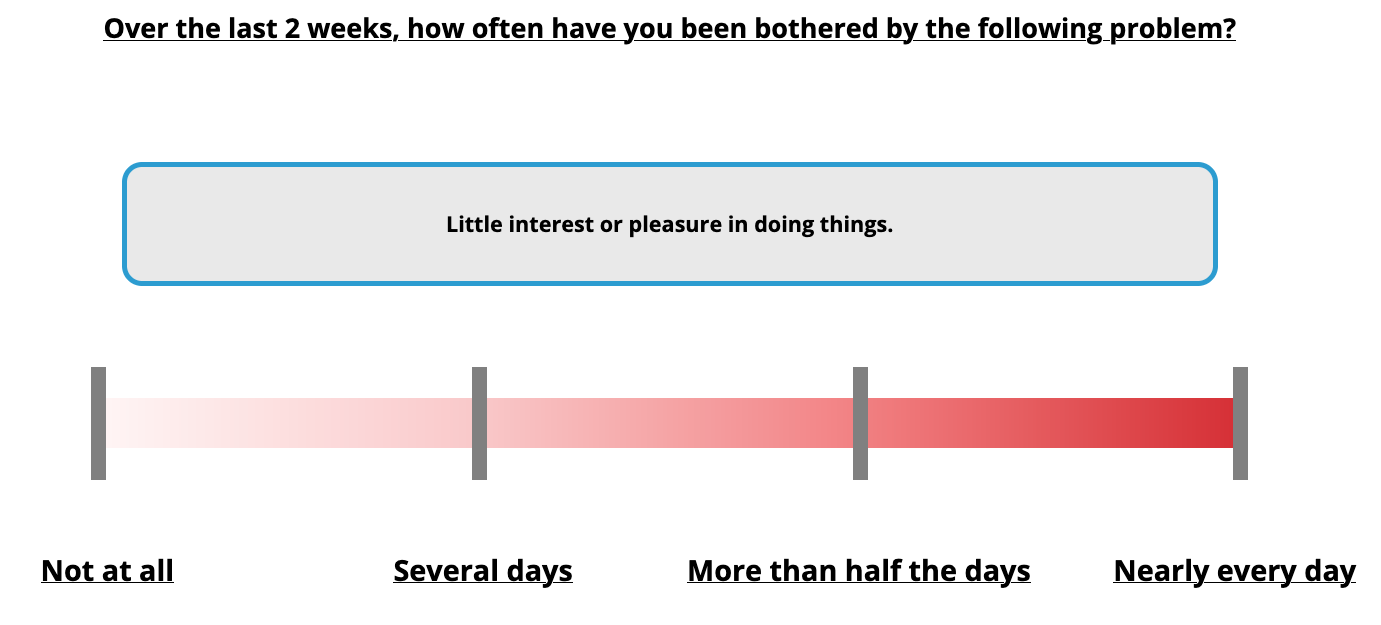}
    \caption{Screenshot of the PHQ-9 rating VAS scale used in Study 3 for example question}
    \label{fig:apdx:vas_phq9}
\end{figure}

Before and after mood induction, participants were asked PHQ-9 items on a visual analogue scale ranging the labels: Not at all, Several days, More than half the days, Nearly every day. See Fig \ref{fig:apdx:vas_phq9}.

% \newpage
\subsubsection{Word recall}
Before the mood induction, participants were asked twice to observe words being displayed one at a time (in random order) and then recall as many as possible in 45 seconds. We used the following list of words:

VIGOROUS, ENERGETIC, LIVELY, EXHAUSTED, TIRED, DRAINED, JOYFUL, DELIGHTED, HAPPY, UNHAPPY, HOPELESS, MISERABLE, GENUINE, WHOLESOME, ETHICAL, CORRUPT, SLOPPY, UNSAFE.

After the mood induction, we asked them to recall the words again. 

\subsection{Intervention selection}\label{apdx:int_sel}
\subsubsection{Identifying mood induction prompts}
To identify mood-induction diaries, we take all participants' open-ended responses (from study 1) and focus on PHQ-9 mood/hopelessness question (Q2). For each open-ended response, we have true subject score and the expected llm-predicted score. We also calculate the absolute difference between subject and llm score. We filter open-ended responses where the absolute difference is $<=0.75$ and where participant wrote at least 30 words. We then select open-ended responses where participants provided score 0 and score 3 (two extremes). We select the most extreme ones and edit the entries to create four entries ranging 33-59 words each (182-192 words total), aiming to maintain the content and severity of the original responses, representative of the average mental state of interest (low/high mood levels).

\clearpage
\subsubsection{Audio generation}
To generate the audio version of these diary entries, we use Hume AI platform \citep{noauthor_home_nodate}, that offers a text-to-speech model (Octave), sensitive to the meaning of words in context. The model additionally allows for control over the level of expressiveness and other prosodic features. In our case, to control for speech prosody and emotional expressiveness across conditions (keeping these as constant as possible), we designed a custom voice with the following voice prompt:
\begin{lstlisting}
A female voice, standard British English (Modern RP). Prosody: Low pitch variation with a flat, monotone-like contour. Pace: Slow and deliberate, approximately 150 words per minute, with natural pauses. Articulation: Clear and precise but with a casual, soft-spoken quality, avoiding harsh sibilance or plosives. Affect: Consistently neutral and emotionally detached.
\end{lstlisting}

We report the transcripts of the interventions below.
\subsubsection{Positive mood induction transcripts}
\begin{enumerate}
    \item Lately, I’ve been really focused on making good memories and enjoying happy moments with my friends and family. I feel joyful, grateful, and full of positive energy. I think it’s because things have been going my way lately — it’s got me feeling excited and motivated.
    \item I was feeling elated today — there was so much positivity around me. I’m surrounded by people I love, and that really makes me happy. Work’s been great too; I have awesome colleagues, and we just wrapped up a big project together. I’m excited for what’s next and ready to keep achieving more.
    \item I’ve been in a happy, light mood and feeling really in control. I’m going out more freely now, with no anxiety, and actually enjoying hanging out with friends. I’ve just been feeling good doing all the regular life stuff. Overall, I’m optimistic, content, and hopeful about what’s ahead.
    \item I’ve had a really positive couple of weeks, with good things happening both at work and at home. I also took some time off to focus on myself and just relax. It’s such a nice feeling — I just feel happy and content.
\end{enumerate}

\subsubsection{Negative mood induction transcripts}
\begin{enumerate}
    \item I’ve been feeling really overwhelmed lately. It’s been tough — I’ve felt lonely and depressed, like something bad is hanging over me. I can’t even remember the last time I didn’t feel this way.
    \item I feel like I’m wasting my life. It’s hard to enjoy anything. Everything just seems pointless, and I don’t know how to change it. I’m so tired of getting rejection after rejection with job applications. I keep comparing myself to others, and it makes me feel pathetic. The world just feels like a cold, evil place.
    \item I just feel too low to do anything. At work, all I can think about is getting home, and once I’m home, I just want to sleep the day away. Most nights I just vape some weed to try and chill, but the feelings don’t really go away — they’re just always there.
    \item My birthday was supposed to be a happy time with family, but I just felt hollow and empty. It was like this wave of sadness hit me, and it hasn’t gone away since. I feel so hopeless, and even the things I used to love don’t bring me any happiness anymore.
\end{enumerate}

\subsection{Instructions and questions}\label{apdx:s3_instr}
We report the exact instructions given to participants when responding in the study.
\subsubsection{Recreate instructions}
\begin{lstlisting}
Next, you will listen to diary entries from the same person.
As you listen, put yourself in their shoes and imagine you are an actor playing their character.
-
You will need to re-enact each entry by typing what you've heard.

Be as detailed as possible, but don't worry about typos or memorising word by word.

Focus on richly capturing the gist from a first-person perspective.
\end{lstlisting}

\subsubsection{Create instructions - non-autobiographical}
\begin{lstlisting}
Keep acting the same character
This time be creative and come up with what their next diary entries could look like.
Be as detailed as possible!

If you've exhausted one train of thought, start another while staying in character.

Keep typing until the time runs out
\end{lstlisting}

% \clearpage
\subsubsection{Create instructions - autobiographical}
\begin{lstlisting}
Now, think about your own life.
Recall situations and events that might have been similar to the ones you've just heard and described.
Imagine you had to write a diary entry to describe how you felt and what you thought about.
What would you write?
Be as detailed as possible!

If you've exhausted one train of thought, start another.

Keep typing until the time runs out.
\end{lstlisting}

\subsubsection{Positive re-evaluation instructions}
These were asked at the end of the study, the goal was to move participants into the positive mental state by positively re-evaluating their baseline open-ended responses.
\begin{lstlisting}
Revisit both your responses from before about your mood, feelings and energy levels. Now, reframe them more positively and rewrite them here. Remember to give examples of situations or experiences that best illustrate these
\end{lstlisting}




\end{appendices}


% \printbibliography

\end{document}